\documentclass[10pt,journal,compsoc]{IEEEtran}

\ifCLASSOPTIONcompsoc
  \usepackage[nocompress]{cite}
\else
  \usepackage{cite}
\fi
\ifCLASSINFOpdf
\else
\fi

\usepackage{amsmath}
\usepackage{amssymb}
\usepackage{xcolor}
\usepackage{comment}
\usepackage{amsmath,amssymb} 
\usepackage{color}
\usepackage{enumitem}
\usepackage{adjustbox}
\usepackage{subcaption}
\usepackage{amsmath}
\usepackage{amssymb}
\usepackage{booktabs}
\usepackage{caption,fixltx2e}
\usepackage[flushleft]{threeparttable}
\usepackage{pifont}
\usepackage{algorithm}
\usepackage{tabu}
\usepackage{algpseudocode}
\usepackage{graphicx}
\usepackage{multirow}
\usepackage{subcaption}
\usepackage{bbm}
\newcommand{\cmark}{\ding{51}}%
\newcommand{\xmark}{\ding{55}}%

\begin{document}
%
\title{Teacher-Student Architecture for Knowledge Learning: A Survey}
%
%
%
%

\author{Chengming Hu,~\IEEEmembership{}
        Xuan Li,~\IEEEmembership{}
        Dan Liu,~\IEEEmembership{}
        Xi Chen,~\IEEEmembership{}
        Ju Wang,~\IEEEmembership{}
        and Xue Liu~\IEEEmembership{Fellow,~IEEE}
\IEEEcompsocitemizethanks{
\IEEEcompsocthanksitem Chengming Hu, Xuan Li, Dan Liu, Xi Chen, and Xue Liu are with the School of Computer Science, McGill University, Montreal, QC, H3A 0G4, Canada \protect \\
E-mail: chengming.hu@mail.mcgill.ca; xuan.li2@mail.mcgill.ca; daniel.liu@mail.mcgill.ca; xi.chen11@mcgill.ca; xueliu@cs.mcgill.ca
\IEEEcompsocthanksitem Ju Wang is with the School of Information Science and Technology, Northwest University, Xi'an, China \protect\\
E-mail: wangju@nwu.edu.cn}
}

\IEEEtitleabstractindextext{
\begin{abstract}
Although Deep Neural Networks (DNNs) have shown a strong capacity to solve large-scale problems in many areas, such DNNs with voluminous parameters are hard to be deployed in a real-time system. 
To tackle this issue, Teacher-Student architectures were first utilized in knowledge distillation, where simple student networks can achieve comparable performance to deep teacher networks. 
Recently, Teacher-Student architectures have been effectively and widely embraced on various knowledge learning objectives, including knowledge distillation, knowledge expansion, knowledge adaption, and multi-task learning. 
With the help of Teacher-Student architectures, current studies are able to achieve multiple knowledge-learning objectives through lightweight and effective student networks.   
Different from the existing knowledge distillation surveys, this survey detailedly discusses Teacher-Student architectures with multiple knowledge learning objectives.
In addition, we systematically introduce the knowledge construction and optimization process during the knowledge learning and then analyze various Teacher-Student architectures and effective learning schemes that have been leveraged to learn representative and robust knowledge. 
This paper also summarizes the latest applications of Teacher-Student architectures based on different purposes (i.e., classification, recognition, and generation). 
Finally, the potential research directions of knowledge learning are investigated on the Teacher-Student architecture design, the quality of knowledge, and the theoretical studies of regression-based learning, respectively.  
With this comprehensive survey, both industry practitioners and the academic community can learn insightful guidelines about Teacher-Student architectures on multiple knowledge learning objectives. 

\end{abstract}

}

\maketitle

\IEEEdisplaynontitleabstractindextext

%
\IEEEpeerreviewmaketitle


\section{Introduction}

\IEEEPARstart{D}{eep} Neural Networks (DNNs) have witnessed many success in several fields, such as Computer Vision~\cite{minaee2021image} (CV), Communication System~\cite{erpek2020deep}, and Natural Language Processing (NLP)~\cite{otter2020survey}, etc. 
Specifically, to satisfy the robust performance in large-scale tasks, DNNs are generally over-parameterized with complex architectures. 
However, such cumbersome models, meanwhile, need a large amount of training time and bring large computational costs, which pose significant challenges to deploying these models on edge devices and in real-time systems.  

To accelerate the training process, Hinton et al.~\cite{hinton2015distilling} first proposed knowledge distillation for training lightweight models to achieve comparable performance to deep models, which is achieved through compressing the informative knowledge from a large and computationally expensive model (i.e., teacher model) to a small and computationally efficient model (i.e., student model).
With such the Teacher-Student architecture, the student model can be trained under the supervision of the teacher model.
During the training of the student model, the student model not only should predict ground truth labels as closely as possible but also should match softened label distributions of the teacher model. 
Consequently, the compressed student model is able to obtain comparable performance to the cumbersome teacher model and is computational-effectively deployed in real-time applications and edge devices.

Teacher-Student architectures have been commonly applied in knowledge distillation for model compression, and some surveys~\cite{gouKnowledgeDistillationSurvey2021, wang2021knowledge, alkhulaifi2021knowledge} summarized the recent progress of various knowledge distillation techniques with Teacher-Student architectures.  
Specifically, Gou et al.~\cite{gouKnowledgeDistillationSurvey2021} presented a comprehensive survey on knowledge distillation from the following perspectives: knowledge types, distillation schemes, and Teacher-Student architectures. 
Wang et al.~\cite{wang2021knowledge} provided a systematic overview and insight into knowledge distillation with Teacher-Student architectures in CV applications. 
Alkhulaifi et al.~\cite{alkhulaifi2021knowledge} summarized multiple distillation metrics to compare the performances of different distillation methods. 
However, these aforementioned surveys do not discuss knowledge construction and optimization during the distillation process, where the knowledge types and optimization objectives are the important factors in providing informative knowledge for student learning. 
Besides, the existing surveys also do not introduce the purposes of knowledge distillation in various application scenarios.

Different from knowledge distillation, Teacher-Student architecture, meanwhile, is effectively and widely embraced on the other knowledge learning objectives, including knowledge expansion, knowledge adaption, and multi-task learning. 
With the help of Teacher-Student architectures, we are able to achieve multiple knowledge-learning objectives through lightweight and effective student networks. 
With the stronger model capacity and difficult training environments, student networks can learn the expanded knowledge from teacher networks, so that students are able to achieve better performance and generalizability over teachers in more complicated tasks~\cite{xie2020self, sohn2020simple, wang2021data}. 
In the knowledge adaption, student networks are trained on one or multiple target domains, with the adapted knowledge of teacher networks trained on source domains~\cite{matiisen2019teacher, li2019bidirectional}.
Besides, multi-task student networks are built to learn more general feature representations under the supervision of multiple specialized teacher networks, so that such general student networks can be effectively employed in multiple tasks~\cite{ghiasi2021multi}.

\begin{figure*}[t]
\centering
\includegraphics[width=\textwidth]{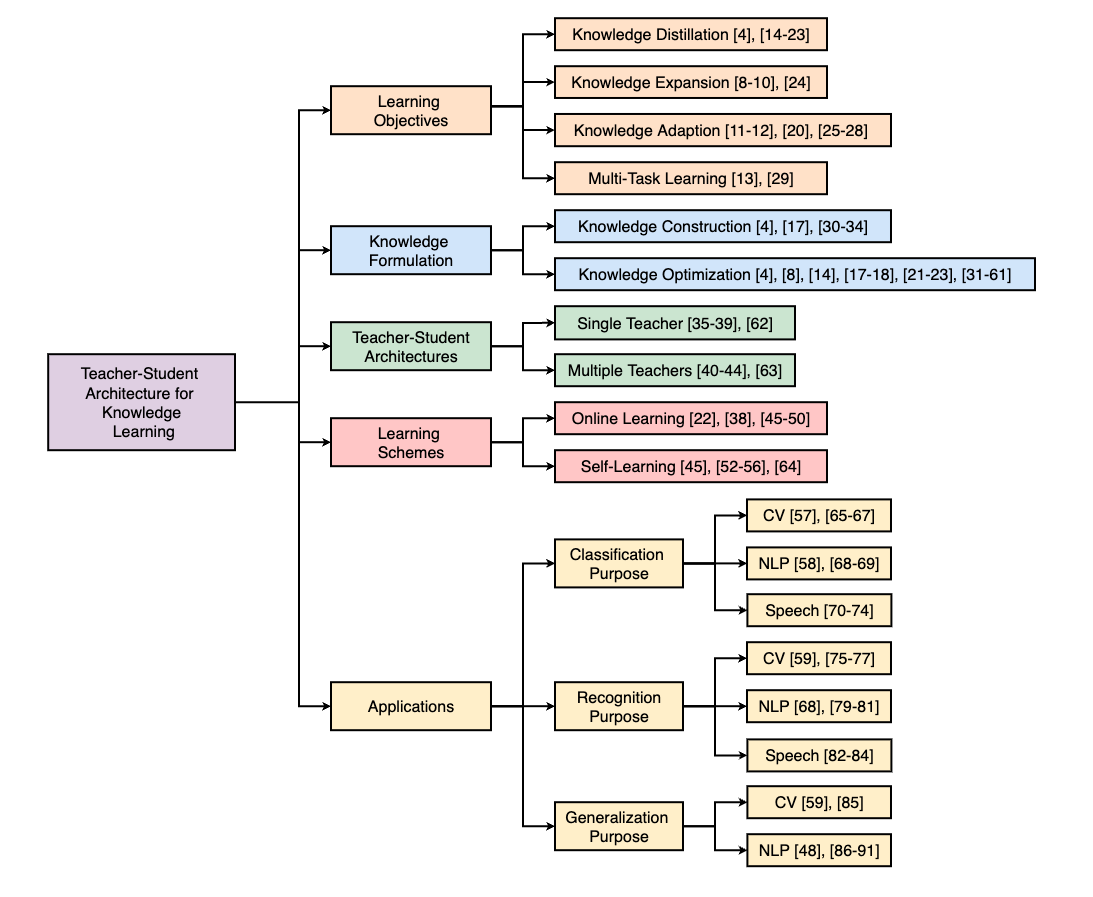}
\centering
\caption{The general taxonomy framework of this survey.}\label{fig:taxonomy}
\end{figure*}

Hence, this survey provides a comprehensive and insightful guideline about Teacher-Student architectures on knowledge learning. 
Different from the existing surveys on knowledge distillation~\cite{gouKnowledgeDistillationSurvey2021, wang2021knowledge, alkhulaifi2021knowledge}, this paper first introduces Teacher-Student architectures on multiple knowledge learning objectives (including knowledge distillation, knowledge expansion, knowledge adaption, and multi-task learning), and then discusses the knowledge construction and optimization processes. 
Moreover, we systematically summarize various Teacher-Student architectures and learning schemes that can be utilized to learn representative and robust knowledge. 
The latest applications of Teacher-Student architectures are also discussed from the perspective of different purposes, including classification, recognition, and generation. 
Finally, we investigate the potential research directions of knowledge learning on the Teacher-Student architecture design, the quality of knowledge, and the theoretical studies of regression-based learning, respectively. 
The overall taxonomy framework of this survey is summarized in Fig.~\ref{fig:taxonomy}.

The \textbf{main contributions} of this survey can be summarized as follows.

\begin{itemize}
    \item Different from the existing surveys on knowledge distillation, we introduce a comprehensive and insightful review of Teacher-Student architectures for multiple knowledge learning objectives, including knowledge distillation, knowledge expansion, knowledge adaption, and multi-task learning.  
    \item We provide a detailed review of knowledge formulation during the knowledge learning process, including knowledge construction and optimization.
    \item We summarize the latest applications of Teacher-Student architectures based on different purposes, including classification, recognition, and generation. 
    \item We discuss the promising research directions on knowledge learning, including the Teacher-Student architecture design, the quality of knowledge, and the theoretical studies of regression-based learning. 
\end{itemize}

The rest of the paper is organized as follows: Section~\ref{sec:learning objective} describes Teacher-Student architectures for different learning objectives, including knowledge distillation, knowledge expansion, knowledge adaption, and multi-task learning. Section~\ref{sec: knowledge formulation} introduces how knowledge is constructed and optimized during the learning process. Section~\ref{sec:architecture} discusses various student networks with single and multiple teacher networks, respectively; and Section~\ref{sec:learning scheme} further describes online learning and self-learning schemes of Teacher-Student networks. Section~\ref{sec:application} summarizes the recent applications of Teacher-Student architectures for different purposes. The future works and conclusions are eventually drawn in Section~\ref{sec:future work} and Section~\ref{sec:conclusion}, respectively.

\section{Learning Objectives}\label{sec:learning objective}

\begin{table}[t]
 \centering
 \caption{Overview of the difference between the 4 different categories. KD is for Knowledge Distillation, KE is for Knowledge Expansion, KA is for Knowledge Adaptation, and MTL is for multi-task learning.}
 \label{tab:speed_comparison}
 \setlength{\tabcolsep}{6pt} 
 \renewcommand{\arraystretch}{1}
\begin{threeparttable}
\begin{tabular}{lcccc}
 \toprule
  & KD & KE & KA & MTL \\
 \midrule
 Size reduction & \cmark & \xmark & \xmark & \xmark \\
 Size increase & \xmark & \cmark & -\tnote{1} & - \\
 Data shift & \xmark & \xmark & \cmark & \xmark \\
 Multi-datasets & \xmark & \xmark & \xmark & \cmark \\
 \bottomrule
\end{tabular}
\begin{tablenotes}
    \item[1] Not the major concern.
\end{tablenotes}
\end{threeparttable}
\end{table}

\subsection{Knowledge Distillation}

\begin{figure}[ht]
\begin{subfigure}{0.45\textwidth}
  \centering
  \includegraphics[width=\textwidth]{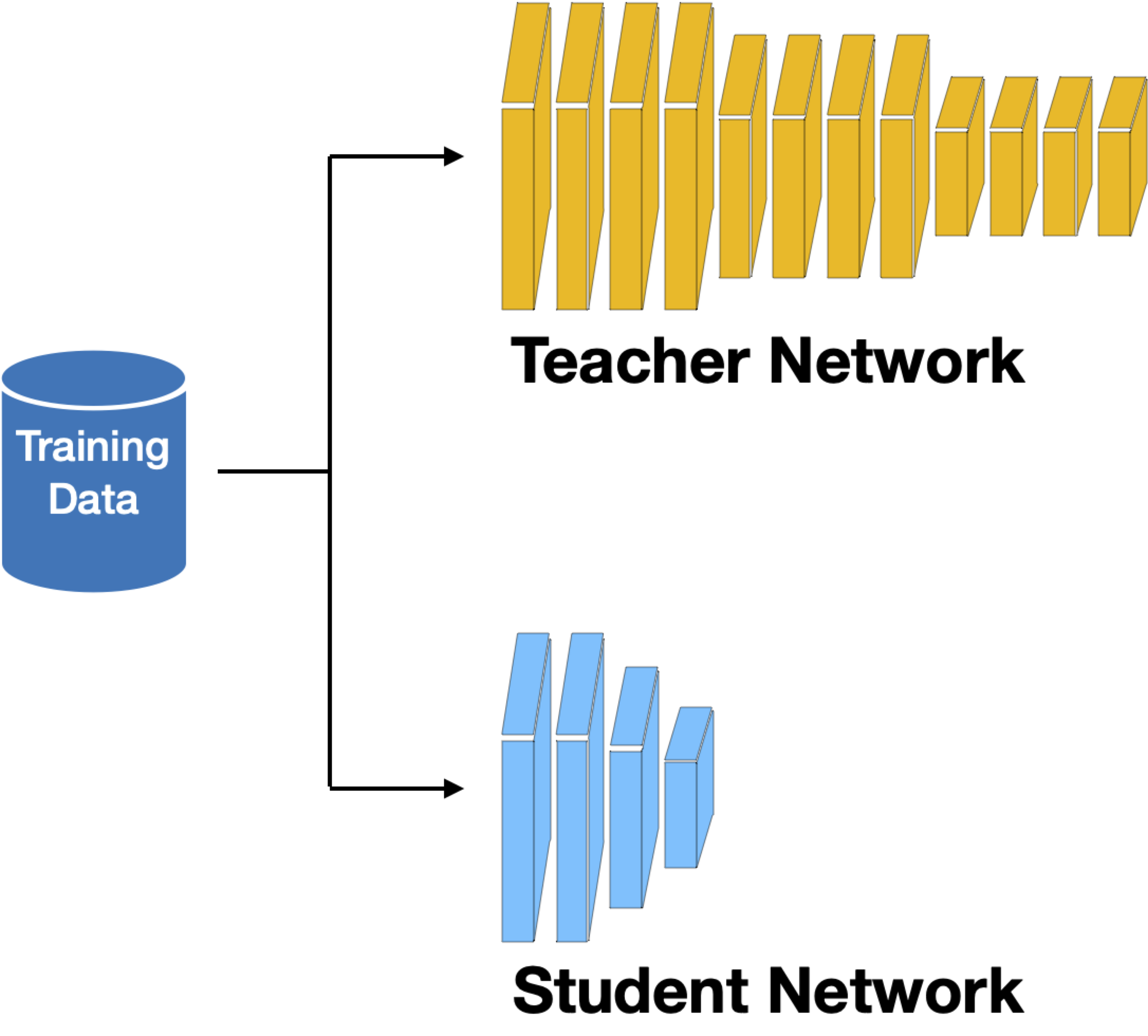}
\end{subfigure}
\caption{Knowledge distillation aims to reduce the size of the student network.}
\label{fig:knowledge_distillation}
\end{figure}

Knowledge distillation focuses on training a student model, using predictions from a larger-sized teacher model. The purpose of knowledge distillation is to have a compact student model while maintaining comparable performance with a teacher model. 

Hinton et al.~\cite{hinton2015distilling} first propose to distill knowledge from multiple models to a single student model for the task of model compression and transfer learning. Tang et al.~\cite{tang2019distilling} compress BERT~\cite{devlin2018bert} to a much light-weight Bi-LSTM~\cite{huang2015bidirectional} for the task of natural language processing. Romero et al.~\cite{romero2014fitnets} suggest the success of deep neural nets is largely attributed to the deep hierarch. Thus they propose to compress wide (a large number of neurons in each layer) and deep teacher models into much narrower (fewer neurons in each layer) and deeper student models. Yim et al.~\cite{yim2017gift}, design the architectures of students and teachers as a $N$-part module, where each module contains various numbers of convolutional layers. Student models generally have a simpler design, and the task for students is to learn each layer output of the teacher.

In a different approach, Wang et al.~\cite{wang2018kdgan} argue that in vanilla \textit{Knowledge Distillation}, it is hard for students to learn all knowledge from teachers, thus students normally show worse performance compared to the teacher. The authors adopt generative adversarial networks~\cite{goodfellow2014generative} to simulate the distillation process. The generator (student model with fewer parameters) learns the distribution of the data, whereas the discriminator (teacher model with more parameters) learns to differentiate if the input is from a student or real.

Tang et al.~\cite{tang2018ranking} propose to distill complicated models in information retrieval or recommendation systems where when a query is given, the model predicts the top K relevant information. Zhang et al.~\cite{zhang2018deep} utilize multiple students to learn from each other during training. The purpose is for performance improvement instead of model compression. Furlane et al.~\cite{furlanello2018born} propose a novel ensemble learning approach, where students and teachers share the same architecture. The $N_{th}$ student is responsible to train the $N+1_{th}$ student. Predictions are averaged in the end.

\subsection{Knowledge Expansion}

\begin{figure}[ht]
\begin{subfigure}{0.45\textwidth}
  \centering
  \includegraphics[width=\textwidth]{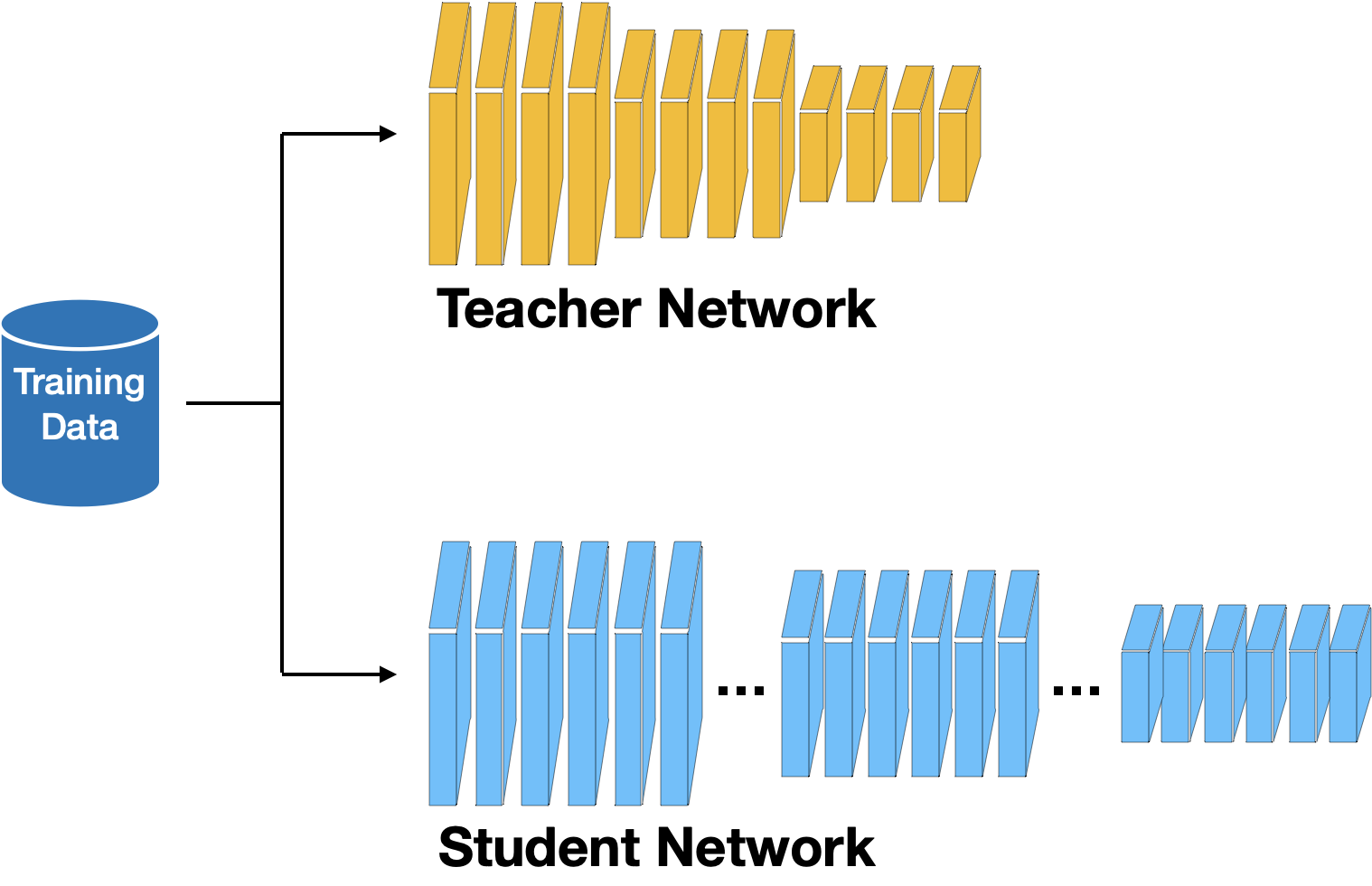}
\end{subfigure}
\caption{Knowledge expansion aims to expand the student network size.}
\label{fig:knowledge_expansion}
\end{figure}

Knowledge expansion differs from knowledge distillation in that, instead of compressing the large teacher model into a smaller student model, it focuses on training a student model with better generalizability and performance than the teacher model. The capacity of the student model is the same or larger than the teacher model.

Xie et al.~\cite{xie2020self} initially propose the concept of Knowledge Expansion, in that they aim to use a teacher model to train a student model with a larger capacity of parameters. They achieve it by initially training the teacher model on ImageNet~\cite{deng2009imagenet}, then they incorporate a privately collected unlabelled dataset, and use the teacher model's prediction as pseudo ground truth to train a larger student model. They demonstrate that with enhanced data augmentation and the additional self-labelled data, the larger student model can outperform the teacher model with an iterative self-training strategy.

In~\cite{sohn2020simple}, the authors adopt a similar strategy for the task of object detection. They use the pre-trained teacher model to generate pseudo labels for the unlabelled images. The student model is trained with strong data augmentation on the pseudo-labelled dataset. Wang et al.~\cite{wang2021data} argue that forcing student models to learn the noisy pseudo labels from teachers may cause noisy label overfitting. They instead propose a curriculum learning strategy, where the student models learn easy samples first, then proceed to hard samples. The criteria for easiness are based on the confidence score from the region proposal network.

\subsection{Knowledge Adaptation}

\begin{figure}[ht]
\begin{subfigure}{0.45\textwidth}
  \centering
  \includegraphics[width=\textwidth]{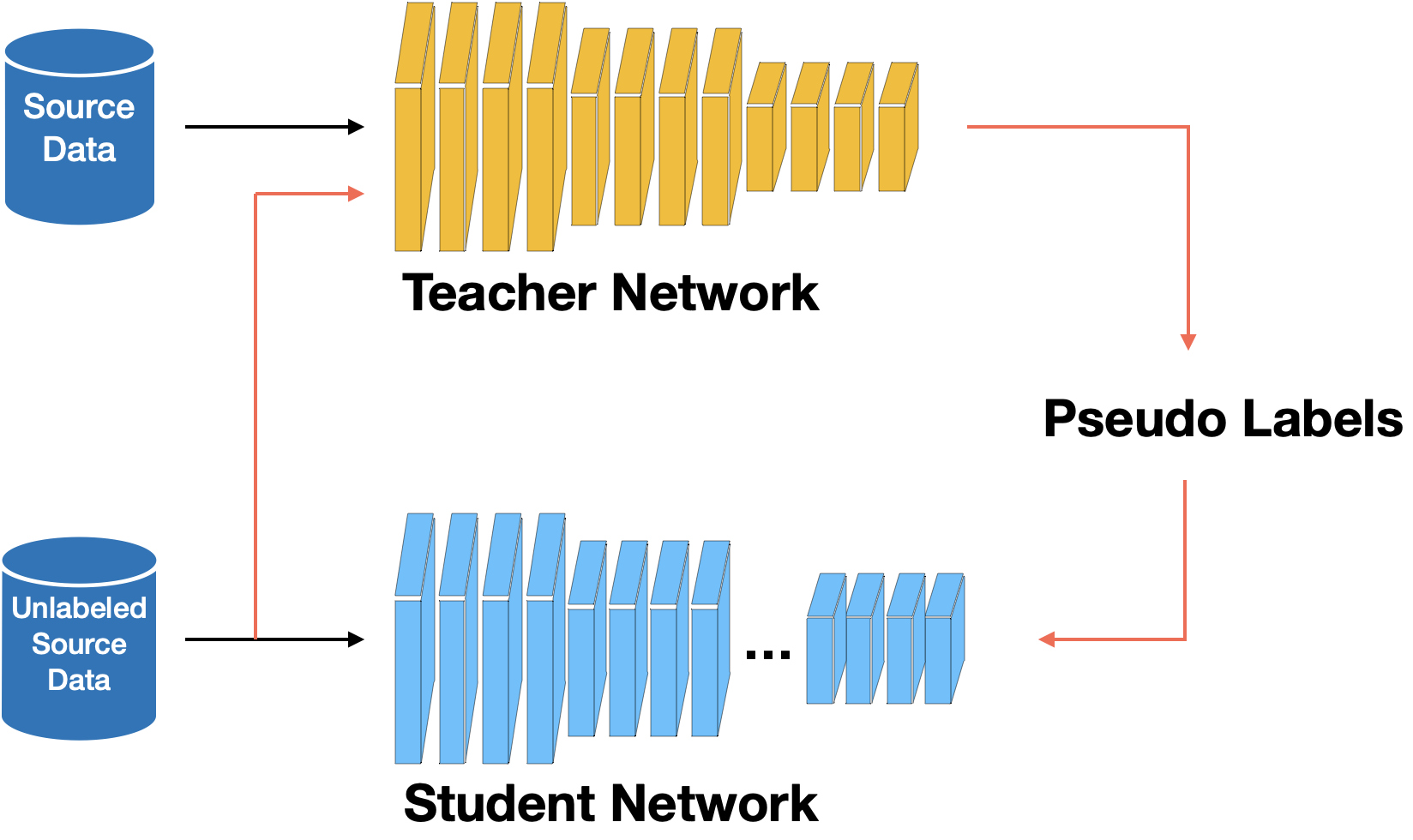}
\end{subfigure}
\label{fig:knowledge_adaptation}
\caption{Knowledge adaptation aims to transfer the knowledge from the source domain to the target domain by training a student network.}
\end{figure}

In~\cite{matiisen2019teacher}, teacher models monitor students' learning progress and decide which task each student should be trained on. In general, a student should be trained on the task where he gains the most performance improvement. But to prevent Catastrophic Forgetting~\cite{kirkpatrick2017overcoming}, students should be trained on tasks where the performance drops.

In~\cite{tsai2018learning}, the authors adopt a novel teacher-student learning scheme by incorporating Generative Adversarial Nets (GANs)~\cite{goodfellow2014generative} into the training. They perform supervised segmentation tasks on the source images with labels. Then they use the trained model to generate predictions for both source and target images. Lastly, they feed both predictions to a teacher model (discriminator). If a teacher model is incapable of differentiating which domain the prediction is from, the trained model has successfully transferred the knowledge from the source domain to the target domain.

In~\cite{hoffman2018cycada}, the authors first propose using CycleGAN~\cite{zhu2017unpaired} to align source domain images with the target domain in an unsupervised approach. Then they further use the aligned source images along with target images to join the adversarial training similar to~\cite{tsai2018learning} where the discriminator is treated as the teacher network.

In~\cite{li2019bidirectional}, the authors extend the work of~\cite{hoffman2018cycada}. They propose a bi-directional learning scheme, where the adversarial training can be in turn used to promote the CycleGAN~\cite{zhu2017unpaired} training. They conduct 3 iterations to obtain the state-of-art performance.

\subsection{Multi-Task Learning}

\begin{figure}[ht]
\begin{subfigure}{0.45\textwidth}
  \centering
  \includegraphics[width=\textwidth]{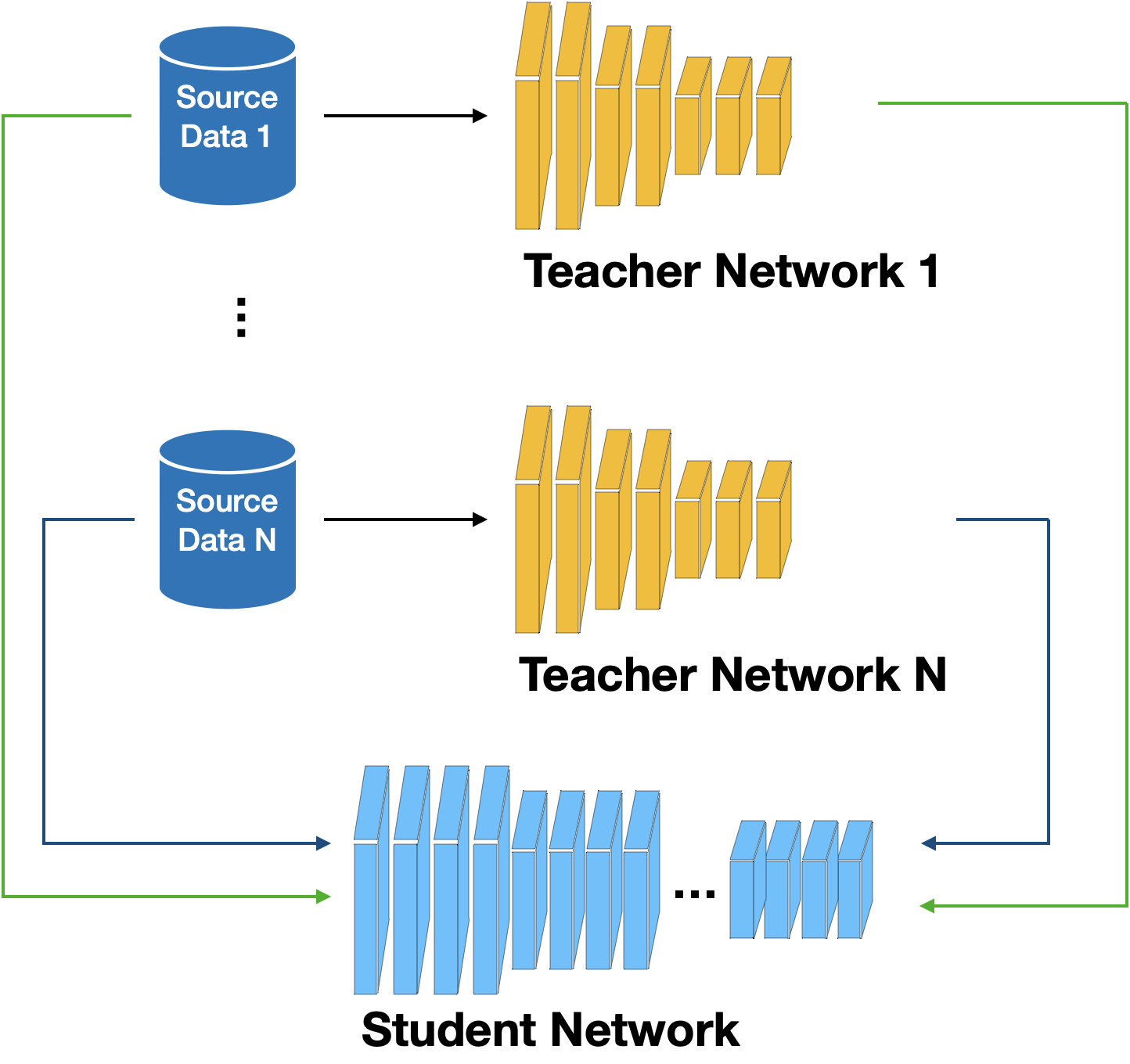}
\end{subfigure}
\label{fig:multi_task_learning}
\caption{Multi-task learning aims to compress knowledge from multiple domains into one domain by training a single student network.}
\end{figure}

Student-teacher training paradigm has been proven effective in various domains. However, adapting to a multi-task context remains challenging. Ghiasi et al~\cite{ghiasi2021multi} propose a multi-task self-training (MuST) strategy which uses multiple independent teachers' models to train one multi-task student model. In particular, they adopt four teacher models, each responsible for classification, detection, segmentation, and depth estimation on four different datasets. After training, these teacher models are used to generate four types of pseudo labels for much larger datasets. The student model is then trained on the dataset with four types of pseudo labels.

In \cite{yang2022cross}, the authors propose a cross-task knowledge distillation framework consisting of three modules: 1) task augmentation: ranking fine-grained cross-tasks through an auxiliary loss. 2) knowledge distillation: sharing ranked knowledge representation across tasks to enforce consistency. 3) Teacher-student training: end-to-end training to improve the generalized representation of knowledge distillation. When training student models using teachers' predictions, it would be detrimental when the predictions are false. To address the issue, the authors propose aligning the predictions with ground truth before sending them to the student model for loss comparison.

\section{Knowledge Formulation}\label{sec: knowledge formulation}


\subsection{Knowledge Construction}

Hinton et al. \cite{hinton2015distilling} propose an approach that compresses ensemble deep networks into a student network with similar depth by forcing the student network to learn to produce softened softmax outputs as well as predict the sample labels. The softened softmax introduces a temperature $\tau$ to represent rich information from the teacher network. This method successfully transfers the generalization ability from a complex teacher network to a small student model.

The work of FitNets~\cite{romero2014fitnets} constructs the knowledge by using a hint layer and the knowledge is the intermediate output of that layer in a teacher network. However,  due to the dimension discrepancy between the intermediate outputs of teacher and student networks, a linear mapping $\beta$ is introduced to unify different dimensions. The squared Euclidean distance between hidden activations is used for knowledge distillation. The authors use a Gram matrix of feature maps from two different layers as knowledge, e.g., the input feature and output feature of a residual block. The distilled knowledge is used as weight initialization for the student network for further optimization. The authors claim that the distilled knowledge can improve the performance of the student network in different tasks.

In attention-based work \cite{zagoruykoPayingMoreAttention2017}, the knowledge is defined by activation and gradient attention maps. The attention maps are defined by the normalized sum of absolute or squared values mapping. The authors show that these attention maps contain valuable information for improving the performance of convolutional neural network architectures.
Kim et al. \cite{kimParaphrasingComplexNetwork2018}, proposes an encoder with a convolutional structure to generate L$_2$ normalized feature embeddings of teacher network as knowledge. The student network is encouraged to produce the same compressed representation to mimic the teacher network.
Heo et al. \cite{heoKnowledgeTransferDistillation2019} uses the pattern of activated outputs as knowledge as a classification problem highly depends on the formation of decision boundaries among classes. The pattern is extracted before the activation function, which keeps much useful information from the teacher network. The distilled knowledge is used for initializing the student network before the classification training.
 
The author uses the L$_2$-normalized outer products of pairwise activations as knowledge based on the observation that semantically similar inputs tend to have similar output patterns. Therefore, the student network will produce similar (dissimilar) output patterns as the trained teacher network when processing the same input pairs. 
The authors of \cite{parkRelationalKnowledgeDistillation2019} use a group of samples to produce feature maps through the teacher network. And take the structural information among those feature maps as knowledge. The structural information, i.e., the distance metric, is defined by the average Euclidean distance and cosine similarity across grouped sample outputs. The teacher and student network are both pre-trained before the knowledge distillation process.
The work of DarkRank \cite{chenDarkrankAcceleratingDeep2018} uses anchor loss in metric learning to formulate the knowledge. The anchor loss helps reduce intra-class distance and enlarge inter-class distance. Such intra/inter-class distance patterns are used as knowledge. The authors also use the embedding layer to encode the hidden activations. The KL divergence and MLE are used to optimize the knowledge distillation process.
Learning from multiple teacher networks shows a hybrid structure to guide the student network. The authors average the softened outputs of multiple teacher networks as one knowledge and apply triplet loss to extract intra/inter-class information as the other knowledge. The voting strategy of multiple teacher networks sums up all of the knowledge distillation loss as the final loss.

\subsection{Knowledge Optimization}
In knowledge optimization, the optimization objective of knowledge distillation usually consists of three parts: regular cross-entropy ($\mathcal{L}_{CE}$) terms, Kullback–Leibler (KL) divergence ($\mathcal{L}_{KL}$) terms, and distance ($\mathcal{L}_{D}$) terms.
The generalized form of the objective function of knowledge distillation is:
\begin{equation}
    \mathcal{L}_{K D}=\mathcal{L}_{CE}\left(\mathbf{y}_{\text {true}}, {P}_{{S}}\right)+\mathcal{L}_{KL}\left({P}_{{T}}^{\tau}, {P}_{{S}}^{\tau}\right)+ \mathcal{L}_{D}
\end{equation}
where $\mathcal{L}_{CE}$ refers to the cross-entropy loss, $\mathcal{L}_{KL}$ refers to Kullback-Leibler divergence, $\mathcal{L}_{D}$ refers to the distance metrics, $\mathbf{y}_{\text {true }}$ denotes the ground truth, ${P}$ denotes 
the feature map or customized vector, e.g., passing through the softmax function $P=\texttt{softmax}(\cdot)$ \cite{chenDarkrankAcceleratingDeep2018}, $_T$ refers to teacher, $_S$ refers to student, and $\tau$ denotes the temperature \cite{romero2014fitnets} of the softmax function.

The generalized distance function that appeared in the objective function is defined as:
\begin{equation}
    \mathcal{L}_{D}=\left\|\mathbf{d}_{l}-\mathbf{d}_{r}\right\|^{P}_N,
\end{equation}
where $\mathbf{d}_{l}$ and $\mathbf{d}_{r}$ denote the vectors to be measured, $^P$ denotes the power, and $_N$ denotes the L$_N$ norm.
For example, in the metric learning scenario, the distance is defined as:
\begin{equation}
\mathcal{L}_{D_\mathcal{M}}
=\left\|\mathbf{a}_{S}-\mathbf{q}_{S}\right\|_{2}^{P},
\end{equation}
where $\mathbf{a}$ denotes the activation, and $\mathbf{q}$ denotes the anchor sample~\cite{chenDarkrankAcceleratingDeep2018}. We summarize the optimization objectives in Table \ref{tab:opt}. ``Cosine'' refers cosine distance.
\begin{table*}[t]
\renewcommand{\arraystretch}{1.10}
\newcommand{\tabincell}[2]{\begin{tabular}{@{}#1@{}}#2\end{tabular}}
\centering
\caption{Optimization Schemes}\label{tab:learning scheme}
\begin{tabular}{|c|c|c|c|}
\hline
\textbf{\tabincell{c}{Works}} & {\tabincell{c}{$\mathcal{L}_{CE}$}} & {\tabincell{c}{$\mathcal{L}_{KL}$}} & {\tabincell{c}{$\mathcal{L}_{D}$}} \\
\hline
\tabincell{c}{Knowledge Distillation \cite{hinton2015distilling}}  & \cmark & \cmark& - \\
\hline
\tabincell{c}{FitNets \cite{romero2014fitnets}}  & \cmark & \cmark& L$_2$ \\
\hline
\tabincell{c}{Noisy Student \cite{xie2020self}}  & \cmark & - &- \\
\hline
\tabincell{c}{Distilling Task-Specific Knowledge from BERT into Simple Neural Networks \cite{tang2019distilling}}  & \cmark & \cmark& L$_2$ \\
\hline
\tabincell{c}{A Gift from Knowledge Distillation\cite{yim2017gift}}  & \cmark & -& L$_2$ \\
\hline
\tabincell{c}{Ranking Distillation\cite{tang2018ranking}}  & - & -& $L_{D_M}$ \\
\hline
\tabincell{c}{Deep Mutual Learning \cite{zhang2018deep}}  & \cmark & \cmark& - \\
\hline
\tabincell{c}{Born-Again Neural Networks\cite{furlanello2018born}}  & \cmark & -& LD \\
\hline
\tabincell{c}{Paraphrasing Complex Network\cite{kimParaphrasingComplexNetwork2018}}  & \cmark & -& LD \\
\hline
\tabincell{c}{Knowledge Transfer via Distillation\cite{heoKnowledgeTransferDistillation2019}}  & \cmark & -& LD \\
\hline
\tabincell{c}{Relational Knowledge Distillation\cite{parkRelationalKnowledgeDistillation2019}}  & \cmark & \cmark& LD \\
\hline
\tabincell{c}{DarkRank\cite{chenDarkrankAcceleratingDeep2018}}  & \cmark & \cmark& LD \\
\hline
\tabincell{c}{BERT Learns to Teach\cite{zhou2021meta}}  & \cmark & \cmark& - \\
\hline
\tabincell{c}{Learning Student-Friendly Teacher Networks for Knowledge Distillation\cite{park2021learning}}  & \cmark & \cmark& - \\
\hline
\tabincell{c}{Knowledge Distillation Meets Self-Supervision\cite{xu2020knowledge}}  & \cmark & \cmark& Cosine \\
\hline
\tabincell{c}{Improved Knowledge Distillation via Teacher Assistant \cite{mirzadeh2020improved}}  & \cmark & \cmark& - \\
\hline
\tabincell{c}{Uninformed Students\cite{bergmann2020uninformed}}  & - & -& $L_2$ \\
\hline
\tabincell{c}{Mean teachers\cite{tarvainen2017mean}}  & \cmark & \cmark& L$_2$ \\
\hline
\tabincell{c}{Adaptive Multi-Teacher\cite{liu2020adaptive}}  & \cmark & \cmark& Cosine \\
\hline
\tabincell{c}{Learning from Multiple Teacher\cite{you2017learning}} & \cmark & \cmark & L$_{D_M}$ \\
\hline
\tabincell{c}{Reinforced Multi-Teacher Selection for Knowledge Distillation\cite{yuan2021reinforced}}  & \cmark & -& - \\
\hline
\tabincell{c}{Knowledge Adaptation: Teaching to Adapt\cite{ruder2017knowledge}}  & \cmark & \cmark& - \\
\hline
\tabincell{c}{Online Knowledge Distillation with Diverse Peers\cite{chen2020online}}  & \cmark & \cmark& $L_2$ \\
\hline
\tabincell{c}{Knowledge Distillation by on-the-Fly\cite{zhu2018knowledge}}  & \cmark & \cmark& - \\
\hline
\tabincell{c}{Online Knowledge Distillation for Efficient Pose Estimation\cite{li2021online}}  & - & \cmark& - \\
\hline
\tabincell{c}{Feature Fusion for Online Mutual Knowledge Distillation\cite{kim2021feature}}  & \cmark & \cmark& - \\
\hline
\tabincell{c}{Attention-based Feature Interaction for Efficient \cite{su2021attention}}  & \cmark & \cmark& - \\
\hline
\tabincell{c}{Peer Collaborative Learning for Online Knowledge distillation\cite{wu2021peer}}  & \cmark & \cmark& - \\
\hline
\tabincell{c}{Large-Scale Domain Adaptation via Teacher-Student Learning\cite{li2017large}}  & \cmark & \cmark& - \\
\hline
\tabincell{c}{Revisiting Knowledge Distillation via Label Smoothing Regularization \cite{yuan2020revisiting}}  & \cmark & \cmark& - \\
\hline

\tabincell{c}{Feature-Map-Level Online Adversarial Knowledge Distillation\cite{chung2020feature}}  & \cmark & \cmark& - \\
\hline
\tabincell{c}{Multi-View Contrastive Learning\cite{yang2021multi}}  & \cmark & \cmark& - \\
\hline
\tabincell{c}{Online Subclass Knowledge Distillation \cite{tzelepi2021online}}  & \cmark & -& L$_2$ \\
\hline
\tabincell{c}{Online Knowledge Distillation via Collaborative Learning \cite{guo2020online}}  & \cmark & \cmark& - \\
\hline

\tabincell{c}{Correlation Congruence for Knowledge Distillation\cite{pengCorrelationCongruenceKnowledge2019}}  & \cmark & \cmark& L$_2$ \\
\hline
\tabincell{c}{Knowledge Distillation from Internal Representations \cite{aguilarKnowledgeDistillationInternal2020}}  & \cmark & \cmark& Cosine \\
\hline
\tabincell{c}{Adversarial Learning of Portable Student Networks \cite{wangAdversarialLearningPortable2018}}  & \cmark & -& L$_2$ \\
\hline
\tabincell{c}{Be Your Own Teacher \cite{zhang2019your}}  & \cmark & \cmark& L$_2$ \\
\hline
\tabincell{c}{Towards Oracle Knowledge Distillation \cite{kang2020towards}}  & \cmark & \cmark& - \\
\hline
\end{tabular}
\label{tab:opt}
\end{table*}

\section{Teacher-Student Architectures}\label{sec:architecture}

\subsection{Single Teacher}

\begin{figure}[t]
\begin{subfigure}{0.48\textwidth}
  \centering
  \includegraphics[width=\textwidth]{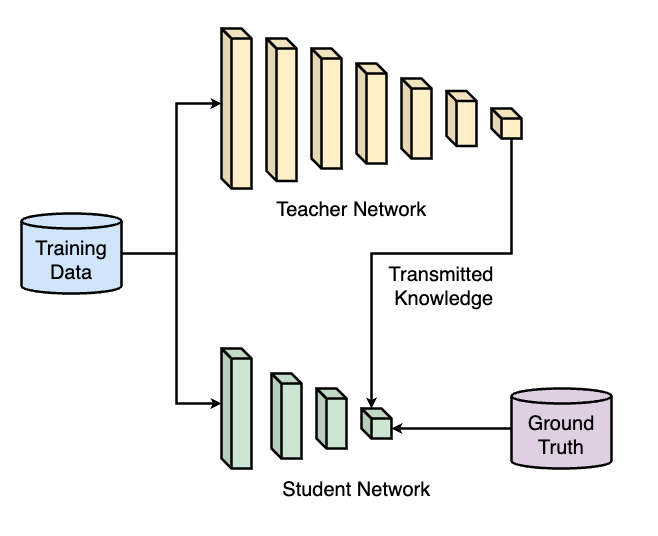}
  \caption{Knowledge learning from single teacher to single student.}
  \label{fig:single teacher}
\end{subfigure}
\begin{subfigure}{0.48\textwidth}
  \centering
  \includegraphics[width=0.95\textwidth]{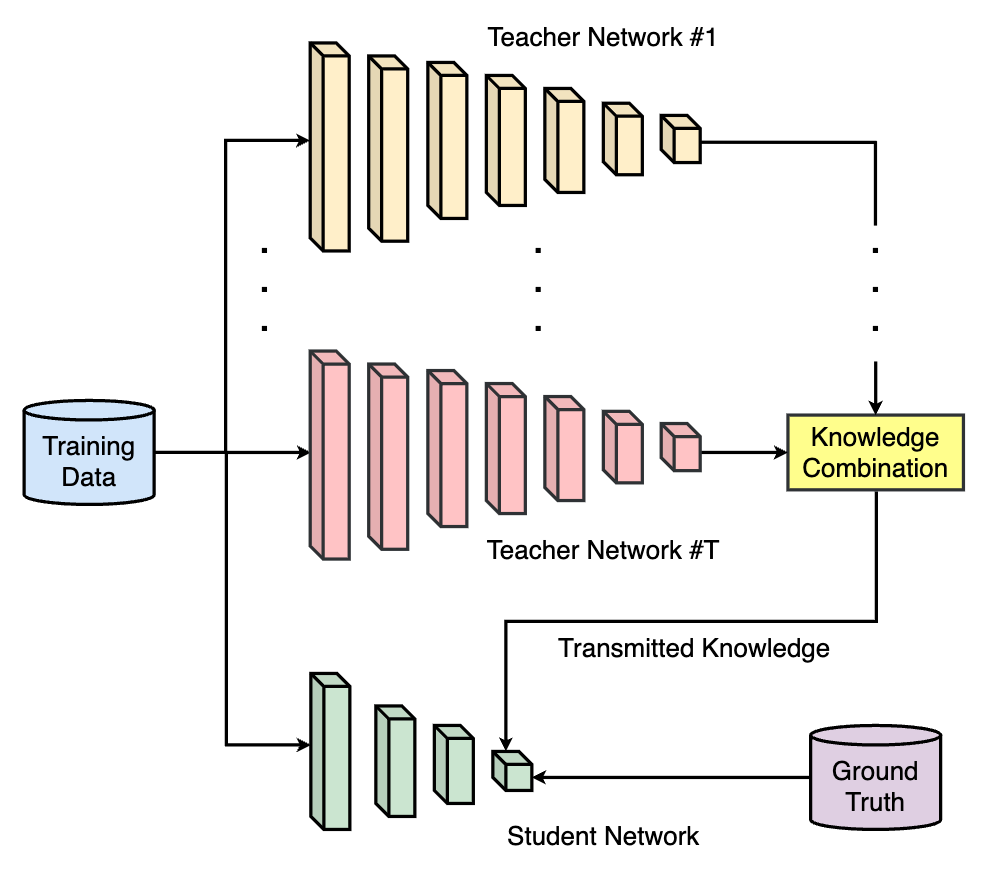}
  \caption{Knowledge learning from multiple teachers to a single student.}
  \label{fig:multiple teachers}
\end{subfigure}
\caption{Teacher-Student architectures.}
\label{fig:architecture}
\end{figure}

Fig.~\ref{fig:architecture} show the architectures of single-single and multiple-single Teacher-Student networks for knowledge learning. 
In the single teacher-based architecture, one student network not only aims to match the ground truth but also learn the transferred knowledge from one teacher network (e.g., soft logits and intermediate feature embeddings). 
MetaDistill~\cite{zhou2021meta} improves the knowledge transmission ability of a teacher network using meta-learning. Specifically, they introduce a pilot update mechanism to formulate the training of a teacher and a student as a bi-level optimization problem, so that the teacher can better transfer knowledge by exploiting feedback about the learning process of students.
SFTN~\cite{park2021learning} first modularizes a teacher and a student network into multiple blocks. The intermediate feature representations from each teacher block are followed by multiple student blocks, respectively, where the teacher and the student are simultaneously trained by minimizing the differences in the feature representations and logits between the teacher and the student. 
To solve the regression problem, Takamoto et al.~\cite{9175560} design and minimize a novel teacher outlier rejection loss function, which aims to discover outliers in training samples using the prediction from a teacher. Moreover, a multi-task network with two outputs, i.e., the estimations of the ground truth and the teacher prediction, is considered to effectively train a student network. 
Xu et al.~\cite{xu2020knowledge} utilize self-supervised learning, when treated as an auxiliary task, to help gain more rounded and dark knowledge from a teacher network. Specifically, contrastive prediction is selected as the self-supervision task to maximize the agreement between a data point and its transformed version via a contrastive loss in latent space. 
Since the large gap between the teacher and the student could degrade the student performance, Mirzadeh et al.~\cite{mirzadeh2020improved} first introduce the teacher assistant (i.e, intermediate-sized network) as a multi-step teacher-student learning to eliminate this gap.
Bergmann et al.~\cite{bergmann2020uninformed} embrace multiple student networks that are simultaneously supervised with a power teacher network pre-trained on a large dataset of natural images, resulting in the accurate pixel-precise anomaly segmentation in high-resolution images. In this way, anomalies can be detected when the students fail to imitate the output of the teacher. 

 \subsection{Multiple Teachers}
Inspired by recent efforts, multiple-teacher networks have
been introduced in knowledge learning, where a student network simultaneously receives knowledge transferred from multiple teachers. Consequently, a student can robustly learn comprehensive and different knowledge under the guidance of multiple teacher networks. 
Averaging multiple teachers is a commonly-used approach to incorporate the potentially diverse knowledge from teachers (i.e., each teacher with an identical importance weight); concretely, a student network aims to learn the average softened logits of multiple teacher networks.
Papernot et al.~\cite{papernot2016semi} introduce multiple teacher networks in Private Aggregation of Teacher Ensembles (PATE), where each teacher network is trained on a private and particular dataset; and then the student network aims to learn the average voting of soft logits of these teachers. 
In Mean Teacher~\cite{tarvainen2017mean}, the average model weights of multiple teachers are regarded as the knowledge to guide the training of the student network. 

Note that multiple teacher networks could be heterogeneous since these teacher networks can be trained in various environments (e.g., different data distributions).  
This suggests that the transferred knowledge from various teachers can contribute differently to the student learning performance, so that the student network may learn more knowledge from similar teacher networks.
As a result, averaging multiple teacher networks could be sub-optimal by assigning each teacher an identical importance weight. 
Thus, to learn more representative and critical knowledge from multiple teacher networks, some advanced teacher weighting approaches are introduced to assign a particular importance weight for each teacher network, so that the student network can learn more representative knowledge from important teachers. 
For example, Adaptive Multi-teacher Multi-level Knowledge Distillation (AMTML-KD)~\cite{liu2020adaptive} includes multiple teacher networks, where each teacher network is learned an instance-level importance weight for adaptively integrating the intermediate feature representations from all teachers. Consequently, a student network can fully learn potentially diverse knowledge from multiple teachers. 
In addition to the logits from multiple teachers, You et al.~\cite{you2017learning} also additionally consider the relative similarity between intermediate representations of samples as one type of dark knowledge to guide the training of a student network. Concretely, the triplets are utilized to encourage the consistency of relative similarity relationships between the student and the teachers.
Yuan et al.~\cite{yuan2021reinforced} formulate the teacher selection problem under an RL framework, where each teacher network is assigned an appropriate importance weight based on various training samples and the outputs of teacher networks. Then, multiple teacher networks are randomly selected based on the learned importance weights to guide the training of the student network at each epoch. 
Ruder et al.~\cite{ruder2017knowledge} consider multi-teacher knowledge distillation for the domain adaptation and design teacher importance weights according to the data similarity between source domains and a target domain. Specifically, each teacher network is trained on a source domain, and the teacher importance weights are increased on the source domains similar to the target domain.

\section{Learning Schemes}\label{sec:learning scheme}

\subsection{Online Learning}

Table~\ref{tab:learning scheme} compares the different learning schemes in terms of teacher and student learning statuses, as well as their role statuses. 
The classic learning scheme is offline learning~\cite{hinton2015distilling}, which represents that student networks learn the knowledge transferred from pre-trained teacher networks. Specifically, powerful teacher networks are first completely well-trained on large-scale datasets, and then transfer the knowledge to guide the training of compact student networks. The roles of teacher and student networks are not exchanged during the student training process. 
It is important to note that offline learning is not included in the scope of this survey paper, and we refer readers to the existing surveys that have provided comprehensive reviews on the offline learning~\cite{gouKnowledgeDistillationSurvey2021, alkhulaifi2021knowledge}. 

Although the offline learning scheme is straightforward and computationally efficient, powerful pre-trained teacher networks and large-scale datasets could be limited~\cite{mirzadeh2020improved, zhang2018deep, chen2020online}. To solve such issues, an online learning scheme is further commonly utilized to simultaneously train student and teacher networks, so that the whole knowledge learning process can be end-to-end trainable.
For instance, in On-the-fly Native Ensemble (ONE) learning strategy~\cite{zhu2018knowledge, li2021online}, a native ensemble teacher network is created from multiple student branches on-the-fly and simultaneously trained with these student branches. With student branches, both teachers and students are more efficient to be trained with superior generalization performance and without asynchronous model updates.
Due to the gating module on the shared layer, student branches are limited to the same network architecture in ONE learning strategy. To overcome this limitation, Feature Fusion Learning (FFL)~\cite{kim2021feature} is further proposed to allow student branches to be applicable to any architecture.
To better boost the knowledge distillation process, Su et al.~\cite{su2021attention} additionally introduce an attention mechanism to capture important and high-level knowledge, so that teachers and students can be dynamically and effectively trained with the help of the valuable knowledge.  
In Peer Collaborative Learning (PCL)~\cite{wu2021peer}, a peer ensemble teacher is trained with the ensemble feature representation of multiple  student peers. Besides, a temporal peer means the teacher is individually built for each student peer to collaboratively transfer knowledge among student peers.

\begin{table}[t]
\renewcommand{\arraystretch}{1.10}
\newcommand{\tabincell}[2]{\begin{tabular}{@{}#1@{}}#2\end{tabular}}
\centering
\caption{Comparison Between Different Learning Schemes}\label{tab:learning scheme}
\begin{tabular}{|c|c|c|c|}
\hline
\textbf{\tabincell{c}{Learning \\ Scheme}} & \textbf{\tabincell{c}{Teacher \\ Training \\ Status}} & \textbf{\tabincell{c}{Student \\ Training \\ Status}} & \textbf{Role Status} \\
\hline
\tabincell{c}{Offline learning \\ \cite{hinton2015distilling, romero2014fitnets, li2017large}}  & \tabincell{c}{Well trained} & \tabincell{c}{To be trained} & Static \\
\hline
\tabincell{c}{Online learning \\ \cite{mirzadeh2020improved, zhang2018deep, chen2020online, zhu2018knowledge} \\ \cite{li2021online, kim2021feature, su2021attention, wu2021peer}} & \tabincell{c}{To be trained} & \tabincell{c}{To be trained} & Static \\ 
\hline
\tabincell{c}{Self-learning \\ \cite{yuan2020revisiting, pang2020self, chung2020feature} \\ \cite{yang2021multi, tzelepi2021online, guo2020online}}  & \tabincell{c}{To be trained} & \tabincell{c}{To be trained} & Dynamic \\
\hline
\end{tabular}
\end{table}

\subsection{Self-Learning}

The self-learning scheme is one particular scheme of online learning. Different from the classic online learning methods, the roles of student and teacher networks are dynamic during the iterative learning process, which indicates that student and teacher networks can be exchanged or student networks are able to learn the knowledge from themselves (i.e., Teacher-free knowledge learning) in the self-learning scheme. 
Yuan et al.~\cite{yuan2020revisiting} demonstrate knowledge distillation as a type of label smoothing regularization, and thus label smoothing regularization can be further regarded as a virtual teacher. Based on these analyses, Teacher-free Knowledge Distillation is further proposed, where a student can learn the knowledge from itself or a manually-designed regularization distribution. 
Pang et al.~\cite{pang2020self} further leverage the iterative self-learning scheme for the end-to-end video anomaly detection, where the student networks (i.e., the anomaly learners) continuously generate the new pseudo labels to replace the previous ones. 
Instead of a powerful teacher network, OKDDip consists of multiple auxiliary student peers and one group leader in the two-level distillation~\cite{chen2020online}. Each student peer is self-trained on the knowledge aggregated from other peers with different importance weights. Then all peers further guide the training of the group leader during the second distillation. 
Chung et al.~\cite{chung2020feature} transfer the feature maps and the soft logits through the online adversarial knowledge distillation, where each network has a discriminator that distinguishes the feature maps from its own and other networks. Through the adversarial training of the discriminator, each network can learn the distribution of feature maps from other networks by fooling the corresponding discriminator. 
Multi-view Contrastive Learning (MCL)~\cite{yang2021multi} captures the correlation of feature representations of multiple student peers, and each peer provides a unique feature representation that suggests a specific view of input data. These feature representations can be regarded as more powerful knowledge to be distilled for the effective training of student peers.
Online Subclass Knowledge Distillation (OSKD)~\cite{tzelepi2021online} reveals the similarities inside each class to capture the shared semantic information among subclasses. During the online distillation process, each sample moves closer to the representations of the same subclass while further away from that of different subclasses. 
Recent online teacher-student learning benefits from collaborative learning, where mutual knowledge is shared to enhance the learning ability of both teachers and students. 
In Knowledge Distillation via Collaborative Learning (KDCL)~\cite{guo2020online}, multiple students with different capacities are assembled to simultaneously provide high-quality soft logits to instruct themselves with significant performance improvement. Specifically, each student is fed with individually-distorted images to reduce variance against perturbations in the input data domain. 
\section{Applications}\label{sec:application}

This section introduces the applications of knowledge learning. Knowledge learning is used in various application fields of deep learning, such as computer vision, natural language processing, model compression, and recommender systems. We classify the applications of knowledge learning according to the network framework, which can be divided into classification, recognition, and generative architecture. The classification architecture refers to the most common use of the output layer to directly classify images. The recognition architecture refers to the basis for extracting key information from images. It includes face recognition, object detection, action recognition, medical image recognition, etc. The generative architecture includes text, speech and image synthesis.
\subsection{Classification Purpose}
\subsubsection{Computer Vision}
Knowledge distillation is used to solve image classification problems \cite{zhuLowresolutionVisualRecognition2019,bagherinezhadLabelRefineryImproving2018,pengCorrelationCongruenceKnowledge2019,liLearningForgetting2017}. For incomplete, ambiguous and redundant image labels, a label refinement model for self-improvement and label augmentation is proposed \cite{bagherinezhadLabelRefineryImproving2018} to learn soft, informative, collective and dynamic for complex image classification Label. Inspired by to knowledge distillation-based low-resolution face recognition, \cite{zhuLowresolutionVisualRecognition2019} proposed deep feature distillation for low-resolution image classification, where the student's output features are matched with the teacher's output features.

\subsubsection{NLP}
Text classification tasks include sentence classification and sentence classification. It contains both general domain and task-specific knowledge distillation. For sentence classification and matching, \cite{tangDistillingTaskspecificKnowledge2019} proposed task-specific knowledge distillation from a BERT teacher model to a bidirectional long short-term memory network. In \cite{sanhDistilbertDistilledVersion2019}, a lightweight student model named DistilBERT with the same general structure as BERT is designed and learned across various tasks in NLP. In \cite{aguilarKnowledgeDistillationInternal2020} propose a simplified student BERT using the internal representation of the large teacher BERT via internal distillation.
\subsubsection{Speech}
For audio classification, \cite{gaoAdversarialFeatureDistillation2019} developed a multi-stage feature distillation method and employed an adversarial learning strategy to optimize knowledge transfer. To improve noise-robust speech recognition, knowledge distillation is used as a speech enhancement tool and an audio-visual multimodal knowledge distillation method is proposed \cite{perezAudiovisualModelDistillation2020}. Knowledge from the teacher model for visual and auditory data is transferred to the student model for audio data. Essentially, this refinement shares cross-modal knowledge between teachers and students \cite{perezAudiovisualModelDistillation2020,albanieEmotionRecognitionSpeech2018,rohedaCrossmodalityDistillationCase2018}. To efficiently detect acoustic events, a quantitative distillation method using both knowledge distillation and quantification is proposed by \cite{shiCompressionAcousticEvent2019}.
\subsection{Recognition Purpose}
\subsubsection{Computer Vision}
One of the applications of knowledge learning is to enhance face recognition and improve the performance of the model from the perspective of efficiency and accuracy.  For example, in \cite{luoFaceModelCompression2016}, knowledge is transferred from the teacher network to the student network of selected information neurons in the top layer of the teacher network. The transfer of knowledge devised a teacher-weighted strategy in which the feature representation of the hint layer was lost to avoid misguiding by the teacher \cite{wangAdversarialLearningPortable2018}. By using the previous student network to initialize the newer networks, a method of recursive knowledge distillation was devised \cite{yanVargfacenetEfficientVariable2019}. Since most face recognition methods are on the test set that the category/identity is unknown to the training set, such as \cite{duongShrinkTeaNetMillionscaleLightweight2019} described the angle loss. As the face recognition standard is usually a measure of the distance between positive and negative sample features.
\subsubsection{NLP}
BERT has given a great boost to natural language processing as a very large and complex multilingual model \cite{devlinBertPretrainingDeep2019}, and it is also a  model that is difficult to train. The training is very time- and resource-consuming. To address this issue, \cite{sunPatientKnowledgeDistillation2019,jiaoTinybertDistillingBert2020,tangDistillingTaskspecificKnowledge2019,wangImprovedKnowledgeDistillation2019} proposed several lightweight BERT variants using knowledge distillation in a model compression manner. \cite{sunPatientKnowledgeDistillation2019} proposed knowledge distillation for BERT Model Compression (BERT-PKD) for sentiment classification, paraphrase similarity matching, natural language inference, and machine reading comprehension. In this knowledge distillation-based method, the feature representation of the labels from the teacher hint layer is passed to the students. To speed up linguistic inference, \cite{jiaoTinybertDistillingBert2020} propose TinyBERT, a two-stage transformer knowledge distillation. 
\subsubsection{Speech}
Knowledge distillation can be used to solve some tough problems in speech recognition \cite{baiLearnSpellingTeachers2019,asamiDomainAdaptationDNN2017,ghorbaniAdvancingMultiaccentedLstmctc2018}. To overcome the problem of overfitting DNN acoustic models when data is rare, knowledge distillation is used as a regularization method to train an adaptive model under the supervision of the source model \cite{asamiDomainAdaptationDNN2017}.  \cite{ghorbaniAdvancingMultiaccentedLstmctc2018} proposed to train an advanced multi-accent student model by extracting knowledge from multiple accent-specific RNN-CTC models with cross-domain knowledge transfer.
\subsection{Generation Purpose}
\subsubsection{Computer Vision}
To improve the accuracy of image classification \cite{chenAdversarialDistillationEfficient2019} proposed a feature map-based knowledge distillation method for GANs. It transfers knowledge from feature maps to students. Using knowledge distillation, \cite{wangAdversarialLearningPortable2018} design a visual interpretation and diagnosis framework for an image classifier, unifying a teacher-student model for interpretation and a deep generative model for diagnosis. 
\subsubsection{NLP}
In the generative model of natural language processing, neural machine translation is an application scenario for KD. However, like the BERT model, good-performing NMT models are often complex and huge. In this context, a number of extended knowledge distillation methods for neural machine translation have been proposed to obtain lightweight NMT models \cite{hahnSelfknow2019,zhouM2KDMultimodelMultilevel2020,kimSequencelevelKnowledgeDistillation2016}. Recently, \cite{zhouM2KDMultimodelMultilevel2020} shows that the better performance of KD-based NAT models depends heavily on their capacity and the data extracted through knowledge transfer. \cite{gordonExplainingSequencelevelKnowledge2019} explains the good properties of sequence-level knowledge distillation from a data augmentation and regularization perspective. For the sequence generation scenario of NMT, efficient word-level knowledge distillation is extended to sequence level \cite{kim2021feature}. The student model that generates sequences mimics the sequence distribution of the teacher. To address the problem of multilingual diversity, \cite{tanEfficientNetRethinkingModel2019} proposes multi-teacher distillation, where multiple individual models dealing with bilingual pairs are teachers and multilingual models are students. In order to improve the performance of machine translation tasks, \cite{weiOnlineDistillingCheckpoints2019} proposes that in online KD, the model that evaluates the best during training is selected as the teacher and updated by any subsequent better model. If the next model does not perform well, the current teacher model will take its place.
\section{Opportunities and Future Works}\label{sec:future work}

In this section, we discuss the opportunities and potential directions for improving the knowledge-learning process from the following perspectives: the architecture design of teacher and student networks, the quality of knowledge, and the theoretical studies of regression-based knowledge learning.

\subsection{Teacher-Student Architecture Design}

To improve the knowledge-learning process, some studies have investigated the relationship between the Teacher-Student architecture design and learning performance. 
The knowledge is commonly represented using soft labels and feature embeddings from teachers, and bigger and more robust teacher networks can be expected to provide more representative and reliable knowledge. Therefore, the classic approach to train more accurate students is to design large-scale teacher networks. 
However, some recent works investigate that the increasing gap (in size) between student and teacher networks could decrease the learning performance~\cite{mirzadeh2020improved, zhang2019your, kang2020towards}. 
Recently, Neural Architecture Search (NAS) has witnessed many successes in automatically designing neural networks in solving some tasks, such as image classification~\cite{liu2018progressive}, NLP~\cite{wang2020textnas}, speech recognition~\cite{mehrotra2020bench}, among others. 
For example, Kang et al.~\cite{kang2020towards} integrate NAS with oracle knowledge distillation to find the optimal student network structures and operations for potentially achieving better performance than teacher networks. 
With the help of NAS, the efficient student network can be found from the fixed and high-performing teacher network with lower computational cost and less network parameters~\cite{bashivan2019teacher}.  
Hence, NAS can be further incorporated with knowledge learning in the future direction, which searches for the best pair of powerful teacher and combat student networks, leading to efficient learning performance.

\subsection{Quality of Knowledge}

Existing works have been commonly dedicated to improving the knowledge-learning process through designing efficient network structures and transferring better feature representations. 
However, there are fewer studies analyzing the amount of informative knowledge that can be potentially utilized and transferred from teachers. 
Through the quantification of visual concepts encoded in teacher networks, Cheng et al.~\cite{cheng2020explaining} explain the success of knowledge distillation from the following three hypotheses: learning more visual concepts, learning various visual concepts, and yielding more stable optimization directions. 
Miles et al.~\cite{miles2021information} integrate the analysis of information theory with knowledge distillation using infinitely divisible kernels, which achieve the computationally-efficient learning process on the cross-model transfer tasks. 
Therefore, the quantification of knowledge can be investigated in the future research direction, which aims to analyze how much important knowledge can be potentially captured before the knowledge-learning process.

\subsection{Theoretical Understanding of Regression-Based Knowledge Learning}

Currently, most teacher-student architectures are employed on classification-based knowledge learning tasks, where intermediate feature embeddings and soft logits can be commonly represented as dark knowledge transferred to student networks. Moreover, more advanced training schemes and architecture designs are introduced to improve the efficiency of the knowledge-learning process. 
However, fewer works focus on the teacher-student architectures on the regression-based knowledge learning tasks, such as communication traffic forecasting, and energy balancing, among others. 
One promising research direction can be investigated in the theoretical studies of regression-based knowledge learning, such as the representation of dark knowledge on the regression problem.
The final prediction outputs of teacher networks are represented as knowledge to be transferred to student networks~\cite{takamoto2020efficient, kang2021data}, and student networks also aim to mimic the extracted feature representations from teacher networks during the knowledge learning process~\cite{xu2022contrastive, saputra2019distilling}. 
With deeper theoretical studies on regression-based knowledge learning, teacher-student architectures will be further effectively employed in practical applications.

\section{Conclusions}\label{sec:conclusion}

Teacher-Student architectures were first proposed in knowledge distillation, which aims to obtain lightweight student networks to achieve comparable performance to deep teacher networks.  
Different from the existing knowledge distillation surveys~\cite{gouKnowledgeDistillationSurvey2021, wang2021knowledge, alkhulaifi2021knowledge}, our survey paper focuses on Teacher-Student architectures for knowledge learning. Specifically, this paper discusses Teacher-Student architectures for various knowledge learning objectives, including knowledge distillation, knowledge expansion, knowledge adaption, and multi-task learning. 
Besides, this paper not only introduces the knowledge construction and optimization processes but also provides detailed overviews of different architectures and learning schemes of Teacher-Student networks. 
The latest applications of Teacher-Student architectures are summarized based on various Teacher-Student network purposes, including classification, recognition, and generation. Finally, this paper investigates the promising research directions of knowledge learning on the Teacher-Student architecture design, the quality of knowledge, and the theoretical studies of regression-based learning, respectively. 
With this comprehensive survey paper, both industry practitioners and the academic community can learn insightful references about Teacher-Student architectures on multiple knowledge-learning objectives.

\bibliographystyle{IEEEtran}
\bibliography{references}

\begin{thebibliography}{100}
\providecommand{\url}[1]{#1}
\csname url@samestyle\endcsname
\providecommand{\newblock}{\relax}
\providecommand{\bibinfo}[2]{#2}
\providecommand{\BIBentrySTDinterwordspacing}{\spaceskip=0pt\relax}
\providecommand{\BIBentryALTinterwordstretchfactor}{4}
\providecommand{\BIBentryALTinterwordspacing}{\spaceskip=\fontdimen2\font plus
\BIBentryALTinterwordstretchfactor\fontdimen3\font minus
  \fontdimen4\font\relax}
\providecommand{\BIBforeignlanguage}[2]{{%
\expandafter\ifx\csname l@#1\endcsname\relax
\typeout{** WARNING: IEEEtran.bst: No hyphenation pattern has been}%
\typeout{** loaded for the language `#1'. Using the pattern for}%
\typeout{** the default language instead.}%
\else
\language=\csname l@#1\endcsname
\fi
#2}}
\providecommand{\BIBdecl}{\relax}
\BIBdecl

\bibitem{minaee2021image}
S.~Minaee, Y.~Y. Boykov, F.~Porikli, A.~J. Plaza, N.~Kehtarnavaz, and
  D.~Terzopoulos, ``Image segmentation using deep learning: A survey,''
  \emph{IEEE transactions on pattern analysis and machine intelligence}, 2021.

\bibitem{erpek2020deep}
T.~Erpek, T.~J. O’Shea, Y.~E. Sagduyu, Y.~Shi, and T.~C. Clancy, ``Deep
  learning for wireless communications,'' in \emph{Development and Analysis of
  Deep Learning Architectures}.\hskip 1em plus 0.5em minus 0.4em\relax
  Springer, 2020, pp. 223--266.

\bibitem{otter2020survey}
D.~W. Otter, J.~R. Medina, and J.~K. Kalita, ``A survey of the usages of deep
  learning for natural language processing,'' \emph{IEEE transactions on neural
  networks and learning systems}, vol.~32, no.~2, pp. 604--624, 2020.

\bibitem{hinton2015distilling}
G.~Hinton, O.~Vinyals, and J.~Dean, ``Distilling the knowledge in a neural
  network,'' \emph{arXiv preprint arXiv:1503.02531}, 2015.

\bibitem{gouKnowledgeDistillationSurvey2021}
J.~Gou, B.~Yu, S.~J. Maybank, and D.~Tao, ``Knowledge {{Distillation}}: {{A
  Survey}},'' in \emph{International {{Journal}} of {{Computer Vision}}}, vol.
  129, Jun. 2021, pp. 1789--1819.

\bibitem{wang2021knowledge}
L.~Wang and K.-J. Yoon, ``Knowledge distillation and student-teacher learning
  for visual intelligence: A review and new outlooks,'' \emph{IEEE Transactions
  on Pattern Analysis and Machine Intelligence}, 2021.

\bibitem{alkhulaifi2021knowledge}
A.~Alkhulaifi, F.~Alsahli, and I.~Ahmad, ``Knowledge distillation in deep
  learning and its applications,'' \emph{PeerJ Computer Science}, vol.~7, p.
  e474, 2021.

\bibitem{xie2020self}
Q.~Xie, M.-T. Luong, E.~Hovy, and Q.~V. Le, ``Self-training with noisy student
  improves imagenet classification,'' in \emph{Proceedings of the IEEE/CVF
  conference on computer vision and pattern recognition}, 2020, pp.
  10\,687--10\,698.

\bibitem{sohn2020simple}
K.~Sohn, Z.~Zhang, C.-L. Li, H.~Zhang, C.-Y. Lee, and T.~Pfister, ``A simple
  semi-supervised learning framework for object detection,'' \emph{arXiv
  preprint arXiv:2005.04757}, 2020.

\bibitem{wang2021data}
Z.~Wang, Y.~Li, Y.~Guo, L.~Fang, and S.~Wang, ``Data-uncertainty guided
  multi-phase learning for semi-supervised object detection,'' in
  \emph{Proceedings of the IEEE/CVF Conference on Computer Vision and Pattern
  Recognition}, 2021, pp. 4568--4577.

\bibitem{matiisen2019teacher}
T.~Matiisen, A.~Oliver, T.~Cohen, and J.~Schulman, ``Teacher--student
  curriculum learning,'' \emph{IEEE transactions on neural networks and
  learning systems}, vol.~31, no.~9, pp. 3732--3740, 2019.

\bibitem{li2019bidirectional}
Y.~Li, L.~Yuan, and N.~Vasconcelos, ``Bidirectional learning for domain
  adaptation of semantic segmentation,'' in \emph{Proceedings of the IEEE/CVF
  Conference on Computer Vision and Pattern Recognition}, 2019, pp. 6936--6945.

\bibitem{ghiasi2021multi}
G.~Ghiasi, B.~Zoph, E.~D. Cubuk, Q.~V. Le, and T.-Y. Lin, ``Multi-task
  self-training for learning general representations,'' in \emph{Proceedings of
  the IEEE/CVF International Conference on Computer Vision}, 2021, pp.
  8856--8865.

\bibitem{tang2019distilling}
R.~Tang, Y.~Lu, L.~Liu, L.~Mou, O.~Vechtomova, and J.~Lin, ``Distilling
  task-specific knowledge from bert into simple neural networks,'' \emph{arXiv
  preprint arXiv:1903.12136}, 2019.

\bibitem{devlin2018bert}
J.~Devlin, M.-W. Chang, K.~Lee, and K.~Toutanova, ``Bert: Pre-training of deep
  bidirectional transformers for language understanding,'' \emph{arXiv preprint
  arXiv:1810.04805}, 2018.

\bibitem{huang2015bidirectional}
Z.~Huang, W.~Xu, and K.~Yu, ``Bidirectional lstm-crf models for sequence
  tagging,'' \emph{arXiv preprint arXiv:1508.01991}, 2015.

\bibitem{romero2014fitnets}
A.~Romero, N.~Ballas, S.~E. Kahou, A.~Chassang, C.~Gatta, and Y.~Bengio,
  ``Fitnets: Hints for thin deep nets,'' \emph{arXiv preprint arXiv:1412.6550},
  2014.

\bibitem{yim2017gift}
J.~Yim, D.~Joo, J.~Bae, and J.~Kim, ``A gift from knowledge distillation: Fast
  optimization, network minimization and transfer learning,'' in
  \emph{Proceedings of the IEEE Conference on Computer Vision and Pattern
  Recognition}, 2017, pp. 4133--4141.

\bibitem{wang2018kdgan}
X.~Wang, R.~Zhang, Y.~Sun, and J.~Qi, ``Kdgan: Knowledge distillation with
  generative adversarial networks.'' in \emph{NeurIPS}, 2018, pp. 783--794.

\bibitem{goodfellow2014generative}
I.~Goodfellow, J.~Pouget-Abadie, M.~Mirza, B.~Xu, D.~Warde-Farley, S.~Ozair,
  A.~Courville, and Y.~Bengio, ``Generative adversarial nets,'' \emph{Advances
  in neural information processing systems}, vol.~27, 2014.

\bibitem{tang2018ranking}
J.~Tang and K.~Wang, ``Ranking distillation: Learning compact ranking models
  with high performance for recommender system,'' in \emph{Proceedings of the
  24th ACM SIGKDD International Conference on Knowledge Discovery \& Data
  Mining}, 2018, pp. 2289--2298.

\bibitem{zhang2018deep}
Y.~Zhang, T.~Xiang, T.~M. Hospedales, and H.~Lu, ``Deep mutual learning,'' in
  \emph{Proceedings of the IEEE Conference on Computer Vision and Pattern
  Recognition}, 2018, pp. 4320--4328.

\bibitem{furlanello2018born}
T.~Furlanello, Z.~Lipton, M.~Tschannen, L.~Itti, and A.~Anandkumar, ``Born
  again neural networks,'' in \emph{International Conference on Machine
  Learning}.\hskip 1em plus 0.5em minus 0.4em\relax PMLR, 2018, pp. 1607--1616.

\bibitem{deng2009imagenet}
J.~Deng, W.~Dong, R.~Socher, L.-J. Li, K.~Li, and L.~Fei-Fei, ``Imagenet: A
  large-scale hierarchical image database,'' in \emph{2009 IEEE conference on
  computer vision and pattern recognition}.\hskip 1em plus 0.5em minus
  0.4em\relax Ieee, 2009, pp. 248--255.

\bibitem{kirkpatrick2017overcoming}
J.~Kirkpatrick, R.~Pascanu, N.~Rabinowitz, J.~Veness, G.~Desjardins, A.~A.
  Rusu, K.~Milan, J.~Quan, T.~Ramalho, A.~Grabska-Barwinska \emph{et~al.},
  ``Overcoming catastrophic forgetting in neural networks,'' \emph{Proceedings
  of the national academy of sciences}, vol. 114, no.~13, pp. 3521--3526, 2017.

\bibitem{tsai2018learning}
Y.-H. Tsai, W.-C. Hung, S.~Schulter, K.~Sohn, M.-H. Yang, and M.~Chandraker,
  ``Learning to adapt structured output space for semantic segmentation,'' in
  \emph{Proceedings of the IEEE conference on computer vision and pattern
  recognition}, 2018, pp. 7472--7481.

\bibitem{hoffman2018cycada}
J.~Hoffman, E.~Tzeng, T.~Park, J.-Y. Zhu, P.~Isola, K.~Saenko, A.~Efros, and
  T.~Darrell, ``Cycada: Cycle-consistent adversarial domain adaptation,'' in
  \emph{International conference on machine learning}.\hskip 1em plus 0.5em
  minus 0.4em\relax PMLR, 2018, pp. 1989--1998.

\bibitem{zhu2017unpaired}
J.-Y. Zhu, T.~Park, P.~Isola, and A.~A. Efros, ``Unpaired image-to-image
  translation using cycle-consistent adversarial networks,'' in
  \emph{Proceedings of the IEEE international conference on computer vision},
  2017, pp. 2223--2232.

\bibitem{yang2022cross}
C.~Yang, J.~Pan, X.~Gao, T.~Jiang, D.~Liu, and G.~Chen, ``Cross-task knowledge
  distillation in multi-task recommendation,'' \emph{arXiv preprint
  arXiv:2202.09852}, 2022.

\bibitem{zagoruykoPayingMoreAttention2017}
S.~Zagoruyko and N.~Komodakis, ``Paying more attention to attention:
  {{Improving}} the performance of convolutional neural networks via attention
  transfer,'' in \emph{{{ICLR}}}, 2017.

\bibitem{kimParaphrasingComplexNetwork2018}
J.~Kim, S.~Park, and N.~Kwak, ``Paraphrasing complex network: {{Network}}
  compression via factor transfer,'' in \emph{{{NeurIPS}}}, 2018.

\bibitem{heoKnowledgeTransferDistillation2019}
B.~Heo, M.~Lee, S.~Yun, and J.~Choi, ``Knowledge transfer via distillation of
  activation boundaries formed by hidden neurons,'' in \emph{{{AAAI}}}, 2019.

\bibitem{parkRelationalKnowledgeDistillation2019}
W.~Park, D.~Kim, Y.~Lu, and M.~Cho, ``{Relational knowledge distillation},'' in
  \emph{{CVPR}}, 2019.

\bibitem{chenDarkrankAcceleratingDeep2018}
Y.~Chen, N.~Wang, and Z.~Zhang, ``Darkrank: {{Accelerating}} deep metric
  learning via cross sample similarities transfer,'' in \emph{{{AAAI}}}, 2018.

\bibitem{zhou2021meta}
W.~Zhou, C.~Xu, and J.~McAuley, ``Bert learns to teach: Knowledge distillation
  with meta learning,'' \emph{arXiv preprint arXiv:2106.04570}, 2021.

\bibitem{park2021learning}
D.~Y. Park, M.-H. Cha, C.~Jeong, D.~Kim, and B.~Han, ``Learning
  student-friendly teacher networks for knowledge distillation,'' \emph{arXiv
  preprint arXiv:2102.07650}, 2021.

\bibitem{xu2020knowledge}
G.~Xu, Z.~Liu, X.~Li, and C.~C. Loy, ``Knowledge distillation meets
  self-supervision,'' in \emph{European Conference on Computer Vision}.\hskip
  1em plus 0.5em minus 0.4em\relax Springer, 2020, pp. 588--604.

\bibitem{mirzadeh2020improved}
S.~I. Mirzadeh, M.~Farajtabar, A.~Li, N.~Levine, A.~Matsukawa, and
  H.~Ghasemzadeh, ``Improved knowledge distillation via teacher assistant,'' in
  \emph{Proceedings of the AAAI Conference on Artificial Intelligence},
  vol.~34, no.~04, 2020, pp. 5191--5198.

\bibitem{bergmann2020uninformed}
P.~Bergmann, M.~Fauser, D.~Sattlegger, and C.~Steger, ``Uninformed students:
  Student-teacher anomaly detection with discriminative latent embeddings,'' in
  \emph{Proceedings of the IEEE/CVF Conference on Computer Vision and Pattern
  Recognition}, 2020, pp. 4183--4192.

\bibitem{tarvainen2017mean}
A.~Tarvainen and H.~Valpola, ``Mean teachers are better role models:
  Weight-averaged consistency targets improve semi-supervised deep learning
  results,'' \emph{Advances in neural information processing systems}, vol.~30,
  2017.

\bibitem{liu2020adaptive}
Y.~Liu, W.~Zhang, and J.~Wang, ``Adaptive multi-teacher multi-level knowledge
  distillation,'' \emph{Neurocomputing}, vol. 415, pp. 106--113, 2020.

\bibitem{you2017learning}
S.~You, C.~Xu, C.~Xu, and D.~Tao, ``Learning from multiple teacher networks,''
  in \emph{Proceedings of the 23rd ACM SIGKDD International Conference on
  Knowledge Discovery and Data Mining}, 2017, pp. 1285--1294.

\bibitem{yuan2021reinforced}
F.~Yuan, L.~Shou, J.~Pei, W.~Lin, M.~Gong, Y.~Fu, and D.~Jiang, ``Reinforced
  multi-teacher selection for knowledge distillation,'' in \emph{Proceedings of
  the AAAI Conference on Artificial Intelligence}, vol.~35, no.~16, 2021, pp.
  14\,284--14\,291.

\bibitem{ruder2017knowledge}
S.~Ruder, P.~Ghaffari, and J.~G. Breslin, ``Knowledge adaptation: Teaching to
  adapt,'' \emph{arXiv preprint arXiv:1702.02052}, 2017.

\bibitem{chen2020online}
D.~Chen, J.-P. Mei, C.~Wang, Y.~Feng, and C.~Chen, ``Online knowledge
  distillation with diverse peers,'' in \emph{Proceedings of the AAAI
  Conference on Artificial Intelligence}, vol.~34, no.~04, 2020, pp.
  3430--3437.

\bibitem{zhu2018knowledge}
X.~Zhu, S.~Gong \emph{et~al.}, ``Knowledge distillation by on-the-fly native
  ensemble,'' \emph{Advances in neural information processing systems},
  vol.~31, 2018.

\bibitem{li2021online}
Z.~Li, J.~Ye, M.~Song, Y.~Huang, and Z.~Pan, ``Online knowledge distillation
  for efficient pose estimation,'' in \emph{Proceedings of the IEEE/CVF
  International Conference on Computer Vision}, 2021, pp. 11\,740--11\,750.

\bibitem{kim2021feature}
J.~Kim, M.~Hyun, I.~Chung, and N.~Kwak, ``Feature fusion for online mutual
  knowledge distillation,'' in \emph{2020 25th International Conference on
  Pattern Recognition (ICPR)}.\hskip 1em plus 0.5em minus 0.4em\relax IEEE,
  2021, pp. 4619--4625.

\bibitem{su2021attention}
T.~Su, Q.~Liang, J.~Zhang, Z.~Yu, G.~Wang, and X.~Liu, ``Attention-based
  feature interaction for efficient online knowledge distillation,'' in
  \emph{2021 IEEE International Conference on Data Mining (ICDM)}.\hskip 1em
  plus 0.5em minus 0.4em\relax IEEE, 2021, pp. 579--588.

\bibitem{wu2021peer}
G.~Wu and S.~Gong, ``Peer collaborative learning for online knowledge
  distillation,'' in \emph{Proceedings of the AAAI Conference on Artificial
  Intelligence}, vol.~35, no.~12, 2021, pp. 10\,302--10\,310.

\bibitem{li2017large}
J.~Li, M.~L. Seltzer, X.~Wang, R.~Zhao, and Y.~Gong, ``Large-scale domain
  adaptation via teacher-student learning,'' \emph{arXiv preprint
  arXiv:1708.05466}, 2017.

\bibitem{yuan2020revisiting}
L.~Yuan, F.~E. Tay, G.~Li, T.~Wang, and J.~Feng, ``Revisiting knowledge
  distillation via label smoothing regularization,'' in \emph{Proceedings of
  the IEEE/CVF Conference on Computer Vision and Pattern Recognition}, 2020,
  pp. 3903--3911.

\bibitem{chung2020feature}
I.~Chung, S.~Park, J.~Kim, and N.~Kwak, ``Feature-map-level online adversarial
  knowledge distillation,'' in \emph{International Conference on Machine
  Learning}.\hskip 1em plus 0.5em minus 0.4em\relax PMLR, 2020, pp. 2006--2015.

\bibitem{yang2021multi}
C.~Yang, Z.~An, and Y.~Xu, ``Multi-view contrastive learning for online
  knowledge distillation,'' in \emph{ICASSP 2021-2021 IEEE International
  Conference on Acoustics, Speech and Signal Processing (ICASSP)}.\hskip 1em
  plus 0.5em minus 0.4em\relax IEEE, 2021, pp. 3750--3754.

\bibitem{tzelepi2021online}
M.~Tzelepi, N.~Passalis, and A.~Tefas, ``Online subclass knowledge
  distillation,'' \emph{Expert Systems with Applications}, vol. 181, p. 115132,
  2021.

\bibitem{guo2020online}
Q.~Guo, X.~Wang, Y.~Wu, Z.~Yu, D.~Liang, X.~Hu, and P.~Luo, ``Online knowledge
  distillation via collaborative learning,'' in \emph{Proceedings of the
  IEEE/CVF Conference on Computer Vision and Pattern Recognition}, 2020, pp.
  11\,020--11\,029.

\bibitem{pengCorrelationCongruenceKnowledge2019}
B.~Peng, X.~Jin, J.~Liu, D.~Li, Y.~Wu, and Y.~Liu, ``Correlation congruence for
  knowledge distillation,'' in \emph{{{ICCV}}}, 2019.

\bibitem{aguilarKnowledgeDistillationInternal2020}
G.~Aguilar, Y.~Ling, Y.~Zhang, B.~Yao, X.~Fan, and E.~Guo, ``{Knowledge
  distillation from internal representations},'' in \emph{{AAAI}}, 2020.

\bibitem{wangAdversarialLearningPortable2018}
Y.~Wang, C.~Xu, C.~Xu, and D.~Tao, ``Adversarial {{Learning}} of {{Portable
  Student Networks}},'' in \emph{Proceedings of the {{AAAI Conference}} on
  {{Artificial Intelligence}}}, vol.~32, Apr. 2018.

\bibitem{zhang2019your}
L.~Zhang, J.~Song, A.~Gao, J.~Chen, C.~Bao, and K.~Ma, ``Be your own teacher:
  Improve the performance of convolutional neural networks via self
  distillation,'' in \emph{Proceedings of the IEEE/CVF International Conference
  on Computer Vision}, 2019, pp. 3713--3722.

\bibitem{kang2020towards}
M.~Kang, J.~Mun, and B.~Han, ``Towards oracle knowledge distillation with
  neural architecture search,'' in \emph{Proceedings of the AAAI Conference on
  Artificial Intelligence}, vol.~34, no.~04, 2020, pp. 4404--4411.

\bibitem{9175560}
M.~Takamoto, Y.~Morishita, and H.~Imaoka, ``An efficient method of training
  small models for regression problems with knowledge distillation,'' in
  \emph{2020 IEEE Conference on Multimedia Information Processing and Retrieval
  (MIPR)}, 2020, pp. 67--72.

\bibitem{papernot2016semi}
N.~Papernot, M.~Abadi, U.~Erlingsson, I.~Goodfellow, and K.~Talwar,
  ``Semi-supervised knowledge transfer for deep learning from private training
  data,'' \emph{arXiv preprint arXiv:1610.05755}, 2016.

\bibitem{pang2020self}
G.~Pang, C.~Yan, C.~Shen, A.~v.~d. Hengel, and X.~Bai, ``Self-trained deep
  ordinal regression for end-to-end video anomaly detection,'' in
  \emph{Proceedings of the IEEE/CVF conference on computer vision and pattern
  recognition}, 2020, pp. 12\,173--12\,182.

\bibitem{zhuLowresolutionVisualRecognition2019}
M.~Zhu, K.~Han, C.~Zhang, J.~Lin, and Y.~Wang, ``Low-resolution {{Visual
  Recognition}} via {{Deep Feature Distillation}},'' in \emph{{{ICASSP}} 2019 -
  2019 {{IEEE International Conference}} on {{Acoustics}}, {{Speech}} and
  {{Signal Processing}} ({{ICASSP}})}.\hskip 1em plus 0.5em minus 0.4em\relax
  {Brighton, United Kingdom}: {IEEE}, May 2019, pp. 3762--3766.

\bibitem{bagherinezhadLabelRefineryImproving2018}
H.~Bagherinezhad, M.~Horton, M.~Rastegari, and A.~Farhadi, ``Label refinery:
  {{Improving}} imagenet classification through label progression,'' 2018.

\bibitem{liLearningForgetting2017}
Z.~Li and D.~Hoiem, ``Learning without forgetting,'' in \emph{{{IEEE TPAMI}}},
  vol.~40, 2017, pp. 2935--2947.

\bibitem{tangDistillingTaskspecificKnowledge2019}
R.~Tang, Y.~Lu, L.~Liu, L.~Mou, O.~Vechtomova, and J.~Lin, ``Distilling
  task-specific knowledge from bert into simple neural networks,'' 2019.

\bibitem{sanhDistilbertDistilledVersion2019}
V.~Sanh, L.~Debut, J.~Chaumond, and T.~Wolf, ``{Distilbert, a distilled version
  of bert: smaller, faster, cheaper and lighter},'' 2019.

\bibitem{gaoAdversarialFeatureDistillation2019}
L.~Gao, H.~Mi, B.~Zhu, D.~Feng, Y.~Li, and Y.~Peng, ``An {{Adversarial Feature
  Distillation Method}} for {{Audio Classification}},'' in \emph{{{IEEE
  Access}}}, vol.~7, 2019, pp. 105\,319--105\,330.

\bibitem{perezAudiovisualModelDistillation2020}
A.~Perez, V.~Sanguineti, P.~Morerio, and V.~Murino, ``Audio-visual model
  distillation using acoustic images,'' in \emph{{{WACV}}}, 2020.

\bibitem{albanieEmotionRecognitionSpeech2018}
S.~Albanie, A.~Nagrani, A.~Vedaldi, and A.~Zisserman, ``Emotion recognition in
  speech using cross-modal transfer in the wild,'' in \emph{{{ACM MM}}}, 2018.

\bibitem{rohedaCrossmodalityDistillationCase2018}
S.~Roheda, B.~Riggan, H.~Krim, and L.~Dai, ``Cross-modality distillation: {{A}}
  case for conditional generative adversarial networks,'' in \emph{{{ICASSP}}},
  2018.

\bibitem{shiCompressionAcousticEvent2019}
B.~Shi, M.~Sun, C.~Kao, V.~Rozgic, S.~Matsoukas, and C.~Wang, ``Compression of
  acoustic event detection models with quantized distillation,'' 2019.

\bibitem{luoFaceModelCompression2016}
P.~Luo, Z.~Zhu, Z.~Liu, X.~Wang, and X.~Tang, ``Face model compression by
  distilling knowledge from neurons,'' in \emph{{{AAAI}}}, 2016.

\bibitem{yanVargfacenetEfficientVariable2019}
M.~Yan, M.~Zhao, Z.~Xu, Q.~Zhang, G.~Wang, and Z.~Su, ``Vargfacenet: {{An}}
  efficient variable group convolutional neural network for lightweight face
  recognition,'' in \emph{{{ICCVW}}}, 2019.

\bibitem{duongShrinkTeaNetMillionscaleLightweight2019}
C.~Duong, K.~Luu, K.~Quach, and N.~Le, ``{{ShrinkTeaNet}}: {{Million-scale}}
  lightweight face recognition via shrinking teacher-student networks,'' 2019.

\bibitem{devlinBertPretrainingDeep2019}
J.~Devlin, M.~Chang, K.~Lee, and K.~Toutanova, ``Bert: {{Pre-training}} of deep
  bidirectional transformers for language understanding,'' in
  \emph{{{NAACL-HLT}}}, 2019.

\bibitem{sunPatientKnowledgeDistillation2019}
S.~Sun, Y.~Cheng, Z.~Gan, and J.~Liu, ``Patient knowledge distillation for bert
  model compression,'' in \emph{{{NEMNLP-IJCNLP}}}, 2019.

\bibitem{jiaoTinybertDistillingBert2020}
X.~Jiao, Y.~Yin, L.~Shang, X.~Jiang, X.~Chen, and L.~Li, ``{Tinybert:
  Distilling bert for natural language understanding},'' in \emph{{EMNLP}},
  2020.

\bibitem{wangImprovedKnowledgeDistillation2019}
M.~Wang, R.~Liu, N.~Hajime, A.~Narishige, H.~Uchida, and T.~Matsunami,
  ``Improved {{Knowledge Distillation}} for {{Training Fast Low Resolution Face
  Recognition Model}},'' in \emph{2019 {{IEEE}}/{{CVF International
  Conference}} on {{Computer Vision Workshop}} ({{ICCVW}})}.\hskip 1em plus
  0.5em minus 0.4em\relax {Seoul, Korea (South)}: {IEEE}, Oct. 2019, pp.
  2655--2661.

\bibitem{baiLearnSpellingTeachers2019}
Y.~Bai, J.~Yi, J.~Tao, Z.~Tian, and Z.~Wen, ``Learn spelling from teachers:
  Transferring knowledge from language models to sequence-to-sequence speech
  recognition,'' 2019.

\bibitem{asamiDomainAdaptationDNN2017}
T.~Asami, R.~Masumura, Y.~Yamaguchi, H.~Masataki, and Y.~Aono, ``Domain
  adaptation of {{DNN}} acoustic models using knowledge distillation,'' in
  \emph{2017 {{IEEE International Conference}} on {{Acoustics}}, {{Speech}} and
  {{Signal Processing}} ({{ICASSP}})}.\hskip 1em plus 0.5em minus 0.4em\relax
  {New Orleans, LA}: {IEEE}, Mar. 2017, pp. 5185--5189.

\bibitem{ghorbaniAdvancingMultiaccentedLstmctc2018}
S.~Ghorbani, A.~Bulut, and J.~Hansen, ``{Advancing multi-accented lstm-ctc
  speech recognition using a domain specific student-teacher learning
  paradigm},'' in \emph{{SLTW}}, 2018.

\bibitem{chenAdversarialDistillationEfficient2019}
X.~Chen, Y.~Zhang, H.~Xu, Z.~Qin, and H.~Zha, ``Adversarial {{Distillation}}
  for {{Efficient Recommendation}} with {{External Knowledge}},'' in
  \emph{{{ACM Transactions}} on {{Information Systems}}}, vol.~37, Jan. 2019,
  pp. 1--28.

\bibitem{hahnSelfknow2019}
S.~Hahn and H.~Choi, ``Self-knowledge distillation in natural language
  processing,'' {R.A.N.L.P.}, Ed., 2019.

\bibitem{zhouM2KDMultimodelMultilevel2020}
P.~Zhou, L.~Mai, J.~Zhang, N.~Xu, Z.~Wu, and L.~Davis, ``{M2KD: Multi-model and
  multi-level knowledge distillation for incremental learning},'' {B.M.V.C.},
  Ed., 2020.

\bibitem{kimSequencelevelKnowledgeDistillation2016}
Y.~Kim, {Rush}, and A.~M, ``Sequence-level knowledge distillation,'' in
  \emph{{{EMNLP}}}, 2016.

\bibitem{gordonExplainingSequencelevelKnowledge2019}
M.~Gordon and K.~Duh, ``{Explaining sequence-level knowledge distillation as
  data-augmentation for neural machine translation},'' 2019.

\bibitem{tanEfficientNetRethinkingModel2019}
M.~Tan and Q.~Le, ``{{EfficientNet}}: {{Rethinking}} model scaling for
  convolutional neural networks,'' in \emph{{{ICML}}}, 2019.

\bibitem{weiOnlineDistillingCheckpoints2019}
H.-R. Wei, S.~Huang, R.~Wang, X.-y. Dai, and J.~Chen, ``Online {{Distilling}}
  from {{Checkpoints}} for {{Neural Machine Translation}},'' in
  \emph{Proceedings of the 2019 {{Conference}} of the {{North}}}.\hskip 1em
  plus 0.5em minus 0.4em\relax {Minneapolis, Minnesota}: {Association for
  Computational Linguistics}, 2019, pp. 1932--1941.

\bibitem{liu2018progressive}
C.~Liu, B.~Zoph, M.~Neumann, J.~Shlens, W.~Hua, L.-J. Li, L.~Fei-Fei,
  A.~Yuille, J.~Huang, and K.~Murphy, ``Progressive neural architecture
  search,'' in \emph{Proceedings of the European conference on computer vision
  (ECCV)}, 2018, pp. 19--34.

\bibitem{wang2020textnas}
Y.~Wang, Y.~Yang, Y.~Chen, J.~Bai, C.~Zhang, G.~Su, X.~Kou, Y.~Tong, M.~Yang,
  and L.~Zhou, ``Textnas: A neural architecture search space tailored for text
  representation,'' in \emph{Proceedings of the AAAI Conference on Artificial
  Intelligence}, vol.~34, no.~05, 2020, pp. 9242--9249.

\bibitem{mehrotra2020bench}
A.~Mehrotra, A.~G.~C. Ramos, S.~Bhattacharya, {\L}.~Dudziak, R.~Vipperla,
  T.~Chau, M.~S. Abdelfattah, S.~Ishtiaq, and N.~D. Lane, ``Nas-bench-asr:
  Reproducible neural architecture search for speech recognition,'' in
  \emph{International Conference on Learning Representations}, 2020.

\bibitem{bashivan2019teacher}
P.~Bashivan, M.~Tensen, and J.~J. DiCarlo, ``Teacher guided architecture
  search,'' in \emph{Proceedings of the IEEE/CVF International Conference on
  Computer Vision}, 2019, pp. 5320--5329.

\bibitem{cheng2020explaining}
X.~Cheng, Z.~Rao, Y.~Chen, and Q.~Zhang, ``Explaining knowledge distillation by
  quantifying the knowledge,'' in \emph{Proceedings of the IEEE/CVF conference
  on computer vision and pattern recognition}, 2020, pp. 12\,925--12\,935.

\bibitem{miles2021information}
R.~Miles, A.~L. Rodr{\'\i}guez, and K.~Mikolajczyk, ``Information theoretic
  representation distillation,'' \emph{arXiv preprint arXiv:2112.00459}, 2021.

\bibitem{takamoto2020efficient}
M.~Takamoto, Y.~Morishita, and H.~Imaoka, ``An efficient method of training
  small models for regression problems with knowledge distillation,'' in
  \emph{2020 IEEE Conference on Multimedia Information Processing and Retrieval
  (MIPR)}.\hskip 1em plus 0.5em minus 0.4em\relax IEEE, 2020, pp. 67--72.

\bibitem{kang2021data}
M.~Kang and S.~Kang, ``Data-free knowledge distillation in neural networks for
  regression,'' \emph{Expert Systems with Applications}, vol. 175, p. 114813,
  2021.

\bibitem{xu2022contrastive}
Q.~Xu, Z.~Chen, M.~Ragab, C.~Wang, M.~Wu, and X.~Li, ``Contrastive adversarial
  knowledge distillation for deep model compression in time-series regression
  tasks,'' \emph{Neurocomputing}, vol. 485, pp. 242--251, 2022.

\bibitem{saputra2019distilling}
M.~R.~U. Saputra, P.~P. De~Gusmao, Y.~Almalioglu, A.~Markham, and N.~Trigoni,
  ``Distilling knowledge from a deep pose regressor network,'' in
  \emph{Proceedings of the IEEE/CVF International Conference on Computer
  Vision}, 2019, pp. 263--272.

\end{thebibliography}

\end{document}